\documentclass[letterpaper, 10 pt, journal, twoside]{IEEEtran}

\usepackage{amsmath,amsfonts}
\usepackage{algorithmic}
\usepackage{array}
\usepackage[caption=false,font=normalsize,labelfont=sf,textfont=sf]{subfig}
\usepackage{textcomp}
\usepackage{stfloats}
\usepackage{url}
\usepackage{verbatim}
\usepackage{graphicx}
\def\BibTeX{{\rm B\kern-.05em{\sc i\kern-.025em b}\kern-.08em
    T\kern-.1667em\lower.7ex\hbox{E}\kern-.125emX}}
\usepackage{balance}

\usepackage{comment}
\usepackage{xcolor}
\usepackage{multirow}
\usepackage{times}
\usepackage{epsfig}
\usepackage{amsmath}
\usepackage{amssymb}
\usepackage{bm}
\usepackage{siunitx}
\usepackage{graphics} 
\usepackage{graphicx}
\usepackage{float}
\usepackage{mathrsfs}
\usepackage{booktabs}
\usepackage{cite}

\begin{document}


\title{Super LiDAR Intensity for Robotic Perception}

\author{Wei Gao, Jie Zhang, Mingle Zhao, Zhiyuan Zhang, Shu Kong, Maani Ghaffari, \\ Dezhen Song, Chengzhong Xu, and Hui Kong %
\vspace{-20pt}
\thanks{Manuscript received: August 14, 2025; Revised: December 7, 2025; Accepted: January 27, 2026. This letter was recommended for publication by Editor M. Vincze upon evaluation of the Associate Editor and Reviewers' comments. This work was supported in part by the Science and Technology Development Fund of Macau SAR under Grants 0046/2021/AGJ/ and 0067/2023/AFJ. \textit{(Corresponding author: Hui Kong.)}}
\thanks{Wei Gao, Jie Zhang, Mingle Zhao, Shu Kong, Chengzhong Xu, and Hui Kong are with the State Key Laboratory of Internet of Things for Smart City (SKL-IOTSC), Faculty of Science and Technology, University of Macau, Macau, China (e-mail: gw.ga0.wei@connect.um.edu.mo; mc25934@connect.um.edu.mo; zhao.mingle@connect.um.edu.mo; skong@um.edu.mo; czxu@um.edu.mo; huikong@um.edu.mo).}
\thanks{Zhiyuan Zhang is with the School of Computing and Information Systems, Singapore Management University, Singapore (e-mail: zhiyuanzhang@smu.edu.sg).}
\thanks{Maani Ghaffari is with the Department of Naval Architecture and Marine Engineering and Department of Robotics, University of Michigan, Ann Arbor, MI, USA (e-mail: maanigj@umich.edu).}
\thanks{Dezhen Song is with the Department of Robotics, Mohamed bin Zayed University of Artificial Intelligence (MBZUAI), Abu Dhabi, UAE (e-mail: dezhen.song@mbzuai.ac.ae).}
\thanks{Digital Object Identifier (DOI): see top of this page.}
}

\markboth{IEEE Robotics and Automation Letters. Preprint Version. Accepted JANUARY, 2026}
{Gao \MakeLowercase{\textit{et al.}}: Super LiDAR Intensity for Robotic Perception} 


\maketitle


\begin{abstract}
Conventionally, human intuition defines {\textit{vision}} as a modality of {\textit{passive}} optical sensing, relying on ambient light to perceive the environment. However, {\textit{active}} optical sensing, which involves emitting and receiving signals, offers unique advantages by capturing both radiometric and geometric properties of the environment, independent of external illumination conditions. This work focuses on advancing {\textit{active}} optical sensing using Light Detection and Ranging (LiDAR), which captures intensity data, enabling the estimation of surface reflectance that remains invariant under varying illumination. Such properties are crucial for robotic perception tasks, including detection, recognition, segmentation, and Simultaneous Localization and Mapping (SLAM). A key challenge with low-cost LiDARs lies in the sparsity of scan data, which limits their broader application. To address this limitation, this work introduces an innovative framework for generating dense LiDAR intensity images from sparse data, leveraging the unique attributes of non-repeating scanning LiDAR (NRS-LiDAR). We tackle critical challenges, including intensity calibration and the transition from static to dynamic scene domains, facilitating the reconstruction of dense intensity images in real-world settings. The key contributions of this work include a comprehensive dataset for LiDAR intensity image densification, a densification network tailored for NRS-LiDAR, and diverse applications such as loop closure and traffic lane detection using the generated dense intensity images. Experimental results validate the efficacy of the proposed approach, which successfully integrates computer vision techniques with LiDAR data processing, enhancing the applicability of low-cost LiDAR systems and establishing a novel paradigm for robotic vision via active optical sensing--{\textit{LiDAR as a Camera}}. The dataset and code are available at: {\url{https://github.com/IMRL/Super-LiDAR-Intensity}}.
\end{abstract}



\section{Introduction}

Recent advancements in LiDAR technology have greatly enhanced its accessibility, driven by reductions in sensor size and cost, thereby expanding its applicability across a wide range of robotic applications. Beyond delivering precise 3D spatial data, LiDAR systems are capable of capturing intensity from objects, acquiring illumination-invariant environmental information, and rich texture to complement the limitations of pure geometric-based LiDAR sensing. 

    \begin{figure}
        \centering
        \includegraphics[width=0.38\textwidth]{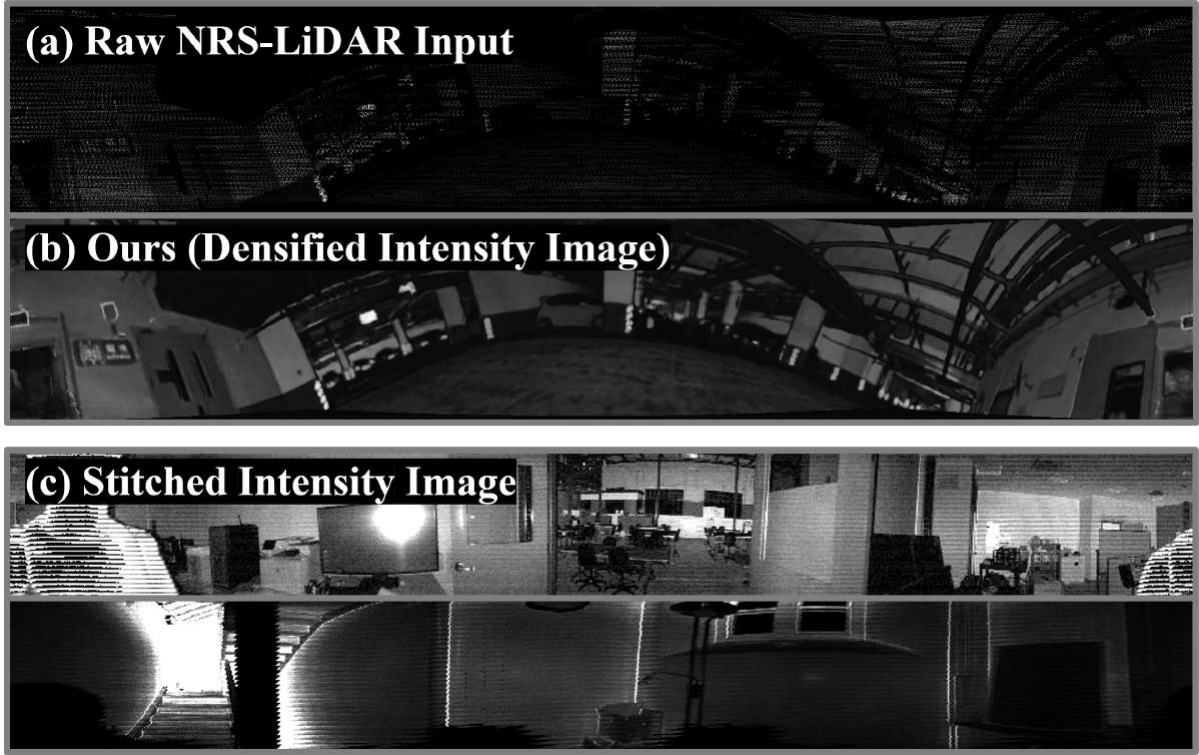}
        \caption{Comparison between the intensity image densified by the proposed method and the true intensity images stitched from high-line LiDAR data. (a) Raw NRS-LiDAR input. (b) Densified 1380 $\times$ 240 intensity image using the proposed method. (c) Stitched 1024 $\times$ 128 intensity images using the high-cost Ouster 128-line LiDAR data from \cite{pfreundschuh2024coin, shan2021robust}. The proposed method effectively eliminates line artifacts and mitigates overexposure in challenging scenarios, generating high-definition, camera-grade intensity images.}
        \label{fig:cover}
        \vspace{-20pt}
    \end{figure}

To date, numerous studies have used LiDAR intensity in various applications, e.g., loop closure detection \cite{shan2021robust,di2021visual}, intensity-enhanced LiDAR odometry \cite{pfreundschuh2024coin, zhang2023ri, du2023real}, segmentation \cite{reymann2015improving, viswanath2024reflectivity}, and detection \cite{hata2014road,li2023LiDAR}. These applications typically depend on dense LiDAR intensity images, which are often generated by high-resolution, costly LiDAR sensors such as Ouster OS1-128 and Velodyne VLS-128. In contrast, low-cost LiDAR sensors, such as Velodyne-16 or Livox MID-360, generally produce sparse data, posing significant challenges for directly applying methods designed for dense intensity images.

Intuitively, generating dense intensity images from sparse LiDAR data holds significant potential for robotic perception, yet this area remains underexplored in existing research \cite{dai2022lidar}. Successfully tackling this challenge would effectively bridge the gap between computer vision and LiDAR sensing techniques, facilitating the adaptation of camera-based methodologies to LiDAR systems. Moreover, it would address the limitations inherent in traditional cameras, such as sensitivity to lighting conditions \cite{gao2024rider,gao2025voyager}, while simultaneously enhancing the texture representation capabilities of LiDAR technology.

\begin{figure}[!t]
        \centering
        \includegraphics[width=0.38\textwidth]{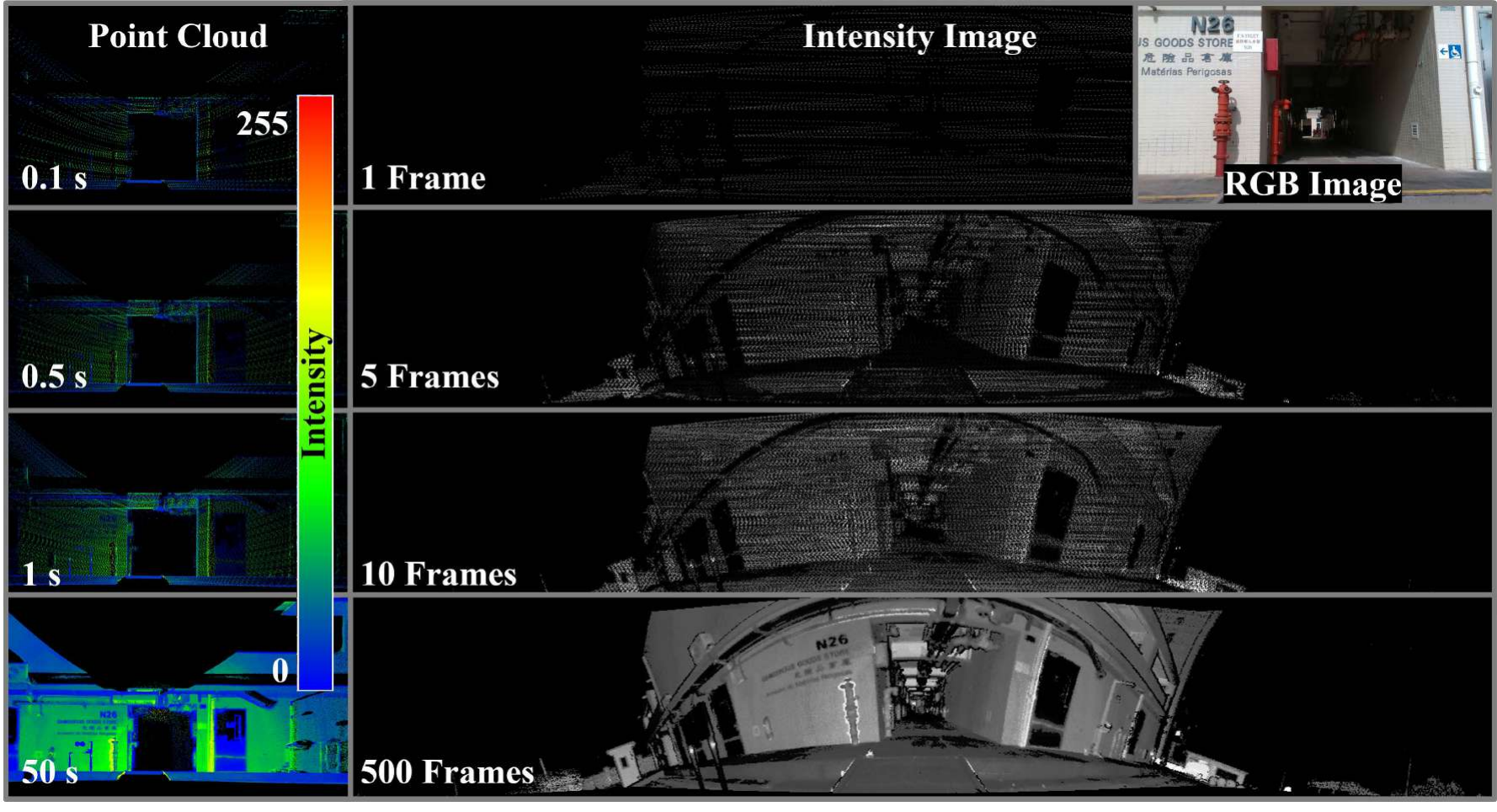}
        \caption{Illustration of the non-repeating scanning mechanism. When stationary, the LiDAR gradually produces denser point clouds over time within the same FoV, allowing the intensity image derived from the point cloud projection to achieve higher density as well. The color of point clouds or gray-scale images represents the intensity value.}
        \label{fig:accumulation_pcl}
        \vspace{-15pt}
    \end{figure}

Creating dense LiDAR intensity images typically depends on high-quality, high-resolution ground-truth data, particularly when using deep neural networks. However, acquiring such data often requires specialized sensors or configurations, posing significant challenges. In this study, we capitalize on the distinctive imaging properties of NRS-LiDAR to produce high-quality ground-truth intensity images (Fig. \ref{fig:cover}). As shown in Fig. \ref{fig:accumulation_pcl}, dense ground-truth data can be acquired in static environments through scan accumulation. Nevertheless, acquiring dense ground-truth data is not feasible when the robot is in motion. In dynamic scenarios, sparse and rapidly changing inputs must be processed in real time, significantly increasing the complexity of dense reconstruction. The disparity between static ground-truth generation and dynamic scene processing underscores the challenges of achieving dense intensity images with low-cost LiDARs. In summary, we face two key challenges: obtaining dense ground-truth for intensity image densification under motion, and utilizing this ground-truth to densify the sparse intensity image. To tackle these issues, we propose a novel approach with the following key contributions:
\begin{enumerate}
    \item The first benchmark dataset for LiDAR intensity densification. Compared to existing datasets \cite{li2021durlar,pfreundschuh2024coin}, the proposed dataset provides paired sparse-to-dense ground-true data, allowing both learning based densification benchmarking and image-level perception tasks.
    
    \item A novel convolutional network for real-time intensity densification for NRS-LiDAR (e.g., Livox MID-360). Our approach effectively addresses two critical challenges: the static-to-motion domain gap and adaptive intensity calibration in LiDAR intensity densification. Compared to existing methods, our method achieves significant improvements in quality and accuracy.
    
    \item The validation in loop closure and traffic lane detection tasks. The proposed method is proven effective under challenging conditions, enabling enhanced practical applications for low-cost LiDAR sensors.
\end{enumerate}

\vspace{-10pt}

\section{Related Work}

\subsection{LiDAR Intensity in Robotics}

In single LiDAR-based applications, numerous studies have primarily focused on exploiting geometric information for robotic perception\cite{qi2017pointnet} and localization\cite{xu2022fast}. Nevertheless, these approaches exhibit limitations in geometrically degenerated environments due to their inability to capture texture information. To address this, researchers have explored fusing camera and LiDAR data to exploit their complementary strengths \cite{wisth2022vilens,lin2022r}. However, this integration presents challenges, including additional hardware, precise calibration, and temporal synchronization. Moreover, visual sensors are inherently limited by illumination conditions.

A promising alternative focuses on leveraging LiDAR's intrinsic intensity properties.
Significant progress has been demonstrated in this domain. A loop closure detection method that synergistically combines geometric and intensity data is proposed \cite{wang2020lidar}. Furthermore, the application of reflectance information has proven particularly effective in robotic terrain analysis, enabling semantic segmentation to distinguish between road surfaces and grasslands\cite{gao2024active}. Although recent learning-based works have advanced intensity simulation by incorporating RGB images and physics-based priors \cite{vacek2021learning,anand2025advancing}, they remain confined to the fidelity of data generation and do not further explore the deployment of such enhanced intensity for downstream robotic tasks.

Recently, dense intensity images, which are generated by high-resolution LiDAR sensors, have been utilized to enhance robotic perception capabilities\cite{zhang2023ri,du2023real}. A notable work is COIN-LIO\cite{pfreundschuh2024coin}, which is a LiDAR-inertial odometry framework integrating LiDAR intensity images with geometric registration to address challenges in geometrically degenerated environments. A hybrid loop closure detection method is introduced in \cite{shan2021robust} by combining LiDAR point cloud and intensity image, achieving superior robustness. However, these advanced methods rely on high-resolution, high-cost LiDAR sensors, making them impractical in resource-constrained scenarios. This study presents a novel approach of reconstructing high-fidelity dense intensity images from sparse ones with low-cost LiDAR, and demonstrates significant potential in real-world real-time robot applications. 

\subsection{Depth and Point Cloud Completion} 

Dense environmental sensing is critical for robotic tasks. However, physical sensors typically provide only sparse measurements. Completing sparse sensory data into dense perceptual representations is thus a crucial research direction. 

Depth completion\cite{uhrig2017sparsity, ma2018sparse} is one of the most extensively studied domains, inspiring us to review dense completion methods. Depth completion utilizes sparse depth data from sensor measurements\cite{qiu2019deeplidar, lu2020depth, yu2021grayscale, liu2022monitored}, multi-view geometry\cite{zuo2021codevio, zhao2022dit}, or data-driven priors\cite{xu2024towards, park2024depth}, to generate dense depth images, enabling continuous and dense perception for navigation and manipulation. Beyond depth images, point cloud completion has also garnered significant attention \cite{zhuang2024survey}. Recent advances employ attention mechanisms \cite{wang2024pointattn}, discrete representations \cite{xiong2023learning}, or unified diffusion models \cite{du2025superpc} to reconstruct 3D shapes from partial observations. However, these methods primarily focus on recovering 3D geometric structures, often neglecting the explicit utilization of intensity information which is crucial for visual perception.

Therefore, directly applying existing depth or point cloud completion approaches to LiDAR intensity images densification is challenging due to the complexities of intensity measurements. Factors such as incidence angle, surface properties, and sensor-to-target distance \cite{fang2014intensity} complicate the learning of latent intensity patterns, especially with limited public training data available. Therefore, developing novel networks specifically for LiDAR intensity images densification is essential. Unlike depth completion, which focuses solely on geometry, LiDAR intensity images densification equips robots with a new, high-resolution eye, thus allowing high-level visual perception to be achieved through LiDAR.


\section{Methodology}
\label{sec: Methodology}

    \begin{figure}
        \centering
        \includegraphics[width=0.38\textwidth]{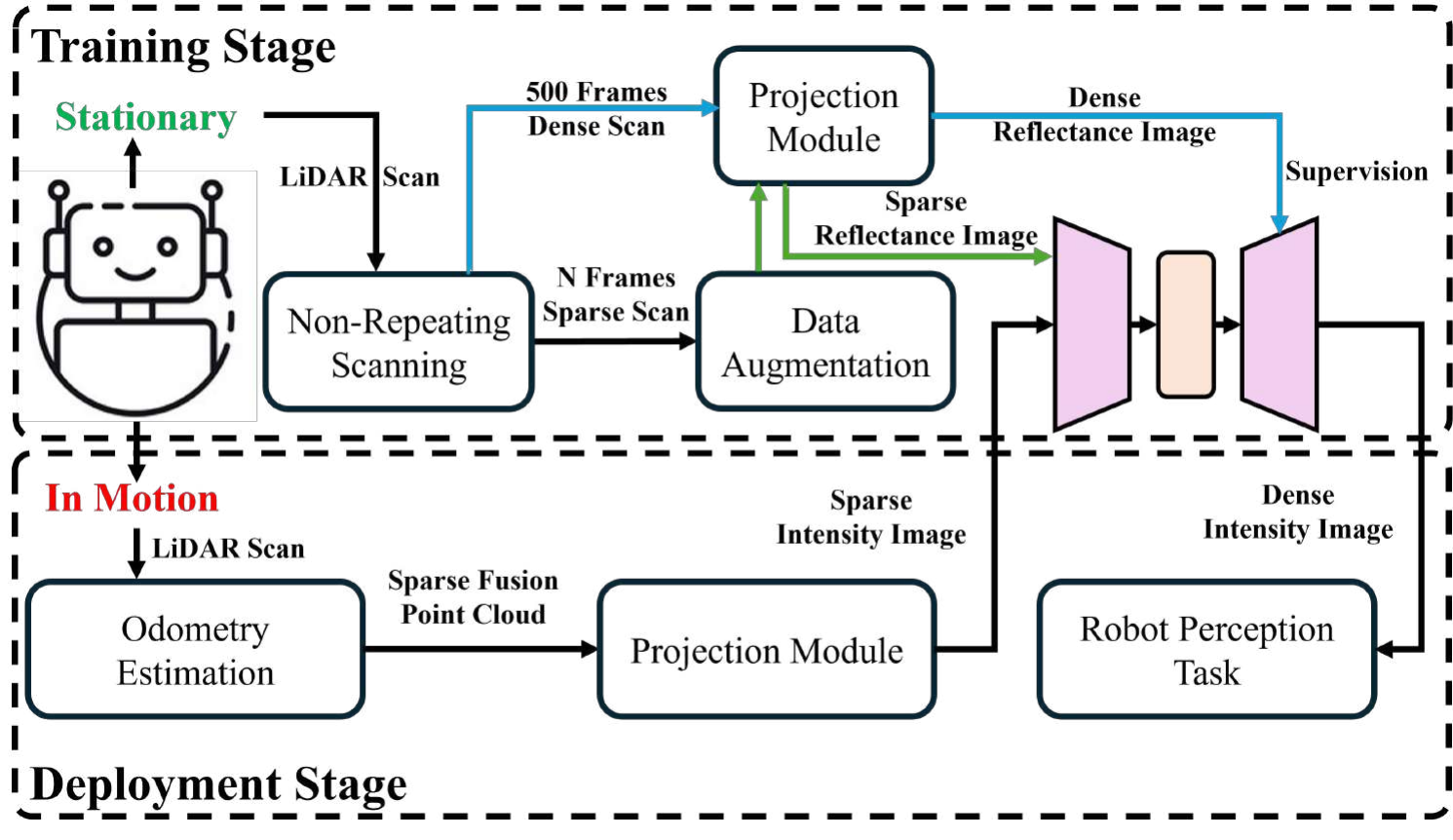}
        \caption{System diagram: During stationarity, the robot utilizes NRS-LiDAR to obtain sparse and dense LiDAR data. Sparse data is augmented to simulate motion characteristics, while dense data serves as supervision to train an intensity image densification network for motion perception tasks.}
        \label{fig:System diagram}
        \vspace{-10pt}
    \end{figure}

Robots in motion often require sufficient time to accumulate data for a comprehensive understanding of their environment. In this section, we detail our proposed method to address this challenge. As shown in Fig. \ref{fig:System diagram}, the purpose of this work is to rely solely on sparse LiDAR data collected by the robot over a short period during motion as input to generate dense and accurate intensity images as output, providing comprehensive environmental information for robotic perception tasks.

\subsection{Dataset Construction}
\label{sec: 3.1}

\subsubsection{Data Acquisition}
The Super LiDAR Intensity dataset was acquired using a Giraffe ground robot platform equipped with a multi-sensor suite: an Intel RealSense D435i camera capturing RGB images at 1280 × 720 resolution, a Livox MID-360 LiDAR sensor with a 360° horizontal and 59° vertical field of view (FoV) operating at 10 Hz point cloud generation rate, and a high-frequency IMU sampling at 200 Hz. Initially, the robot autonomously navigated through selected representative environments. Upon reaching predetermined locations, it maintained a stationary pose for 60 seconds to facilitate the accumulation of multiple LiDAR scans. The synchronized sensor data, including LiDAR point clouds, camera images, and IMU measurements, were recorded as raw ROS (Robot Operating System) packets and systematically organized into dataset collections. To enhance data diversity and minimize inter-dataset correlation, the robot's orientation and position were deliberately adjusted between recording sessions. For the same scene, the robot was either moved a distance \(d > 30 \, \mathrm{m}\) or rotated by an angle \(\theta > 90^\circ\) to capture diverse perspectives and ensure sample uniqueness. The dataset comprises 20 distinct urban scenes, including gardens, libraries, plazas, etc., with an average density of approximately 40 samples per kilometer. To eliminate data leakage, the dataset was split by scenes, ensuring that the training, validation, and test sets were collected from entirely separate environments with no overlap.

\subsubsection{Ground Truth Generation}
This section exploits two projection methods to generate ground-true LiDAR intensity image: panoramic spherical projection and virtual-camera perspective projection.
In the former one, LiDAR points are directly mapped onto a spherical surface. The azimuth \(\theta = \arctan2(y, x)\) and elevation \(\phi = \arctan2\left(z, \sqrt{x^2 + y^2}\right)\) of a specific 3D point are converted into 2D coordinates: \(\text{col} = \left\lfloor\frac{-\theta + 180^\circ}{360^\circ} \cdot W\right\rfloor\) and \(\text{row} = \left\lfloor\frac{\phi_{\text{max}} - \phi}{\phi_{\text{max}} - \phi_{\text{min}}} \cdot H\right\rfloor\)
where \(W\) and \(H\) are the width and height of the generated image. \([\phi_{\text{min}}, \phi_{\text{max}}]\) defines the vertical FoV. The intensity \(I\) is assigned to the corresponding pixel \((\text{row}, \text{col})\). This method aligns with the LiDAR sensor’s FoV, producing images with a resolution of 1380 × 240 pixels.

In the virtual-camera perspective projection, four virtual cameras are placed around the LiDAR sensor to project the point cloud into four views. Using a pinhole camera model, 3D points \((x, y, z)\) are mapped onto a 2D image plane:
    \begin{align}
    \text{col} &= \left\lfloor -f_x \frac{y}{x} + c_x \right\rfloor, \\
    \text{row} &= \left\lfloor H - 1 - \frac{\arctan2(z, x) - \phi_{\text{min}}}{\phi_{\text{max}} - \phi_{\text{min}}} \cdot H \right\rfloor,
    \end{align}
where \(f_x = \frac{W}{2 \cdot \tan\left(\frac{\text{horizontal FoV}}{2}\right)}\), \(f_y = \frac{H}{\phi_{\text{max}} - \phi_{\text{min}}}\), and \(c_x = \frac{W}{2}\). Each view generates intensity images in 240 × 480 pixels. 

\begin{figure} [t]
        \centering
        \includegraphics[width=0.38\textwidth]{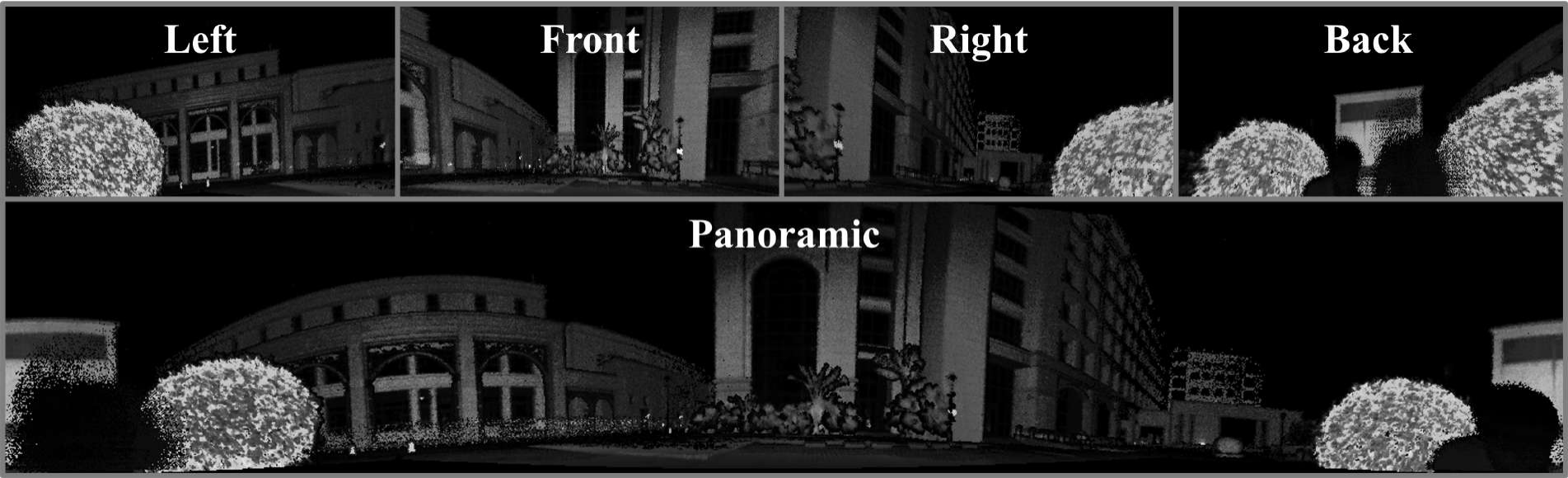}
        \caption{Intensity images obtained by virtual-camera projection mode (top) and panoramic mode (bottom), respectively.}
        \label{fig: four_view}
        \vspace{-10pt}
    \end{figure}

The exemplar intensity images generated by these two projection modes are shown in Fig. \ref{fig: four_view}. Since the density of LiDAR intensity image varies with the number of accumulated scans, we use intensity images accumulated over 500 scans as ground-truth to provide supervision during the training process. Through the aforementioned data acquisition, we constructed the Super LiDAR Intensity dataset, which includes 1,000 raw ROS packets (600K frames) and an additional 5,000 sets of intensity images ranging from sparse (1 scan) to dense (500 scans) configuration, covering diverse urban environments. Additionally, we include specialized subsets: 1) \textit{DCM Supervision} Data for training the compensation module; 2) \textit{Compensation Verification Data}; and 3) \textit{Application Data} targeting specific tasks. The application subset includes data for loop closure detection, with data from three scenarios at two time intervals of the day (morning, evening), using panoramic spherical projection for robust matching across different viewpoints. The other subset is designed for lane detection under varying traffic scenarios during the day and night, where the virtual-camera perspective projection is used (more suitable for this task).

\subsection{Static-to-Dynamic Domain Gap}
\label{sec: 3.2}

The above ground-truth supervision data collection is collected when the LiDAR is stationary. However, the proposed densification method is generally intended for robot perception tasks when the LiDAR is in-motion. 
In practical motion mode, it is necessary to fuse via odometry estimation a certain number of consecutive scans to construct a relatively dense intensity image as network input. When fusing, using too few scans (e.g., only one or two scans) often results in excessively sparse data, leading to lacking sufficient information. On the other hand, fusing too many scans can introduce errors caused by odometry estimation inaccuracies or noise points generated by dynamic objects in the scene. To address the influence of dynamic objects in such scenarios, we adopt the preprocessing method proposed in \cite{wu2024moving} to filter the raw point cloud data, effectively removes motion artifacts caused by dynamic objects and mitigating their adverse effects on the fusion process. In our case, the network input is generated by fusing $N$ consecutive scans using odometry estimates obtained by \cite{xu2022fast}. We empirically selected $N=5$ to balance input density against motion-induced artifacts (see Supplementary Material). By comparing with the static ground-truth training data generation, there obviously exists a gap between the static training data and the dynamic inference ones, which can degrade performance during motion scenarios.

\begin{figure}
        \centering
        \includegraphics[width=0.38\textwidth]{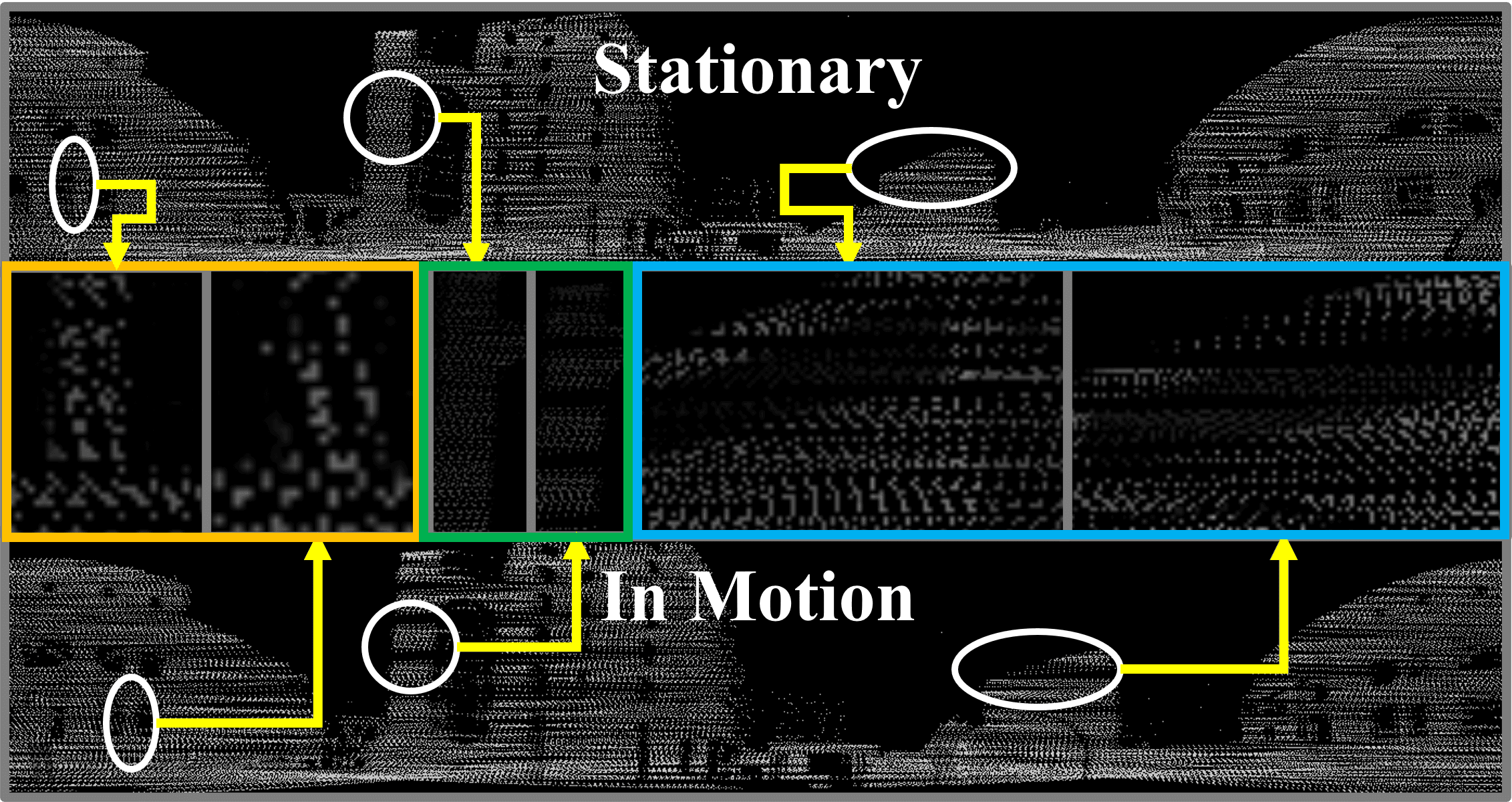}
        \caption{
        Comparison of intensity images fused from 5 scans in stationary and in-motion cases. The image shows the corresponding intensity images. In the stationary case, the 5-scan inputs are aligned, whereas the in-motion case reveals irregularities and misalignment due to motions.}
        \label{static_dynamic}
        \vspace{-10pt}
    \end{figure}

As shown in Fig. \ref{static_dynamic}, the key differences can be described in two aspects. \textit{geometric misalignment:} Odometry-based scan fusion during motion introduces alignment errors due to cumulative motion estimation inaccuracies, leading to inconsistencies in the input. \textit{Motion-induced distribution shift:} Robot motion alters the data distribution, leading to variations in point density, spatial coverage, and alignment compared to the static state. Specifically, motion reduces the overlap between scans, causing certain regions to become sparse within the LiDAR's field of view, further complicating adaptation to dynamic conditions.

The input in the static training setup can be represented as:
    \begin{equation}
    \mathbf{X}_{\text{train}} = \mathcal{P} \left( \sum_{i=1}^{N} \mathbf{R}_i(x, y, z) \right),
    \end{equation}
where \(\mathcal{P}( \cdot )\) represents the projection operation from 3D point cloud to 2D intensity image, \(\mathbf{R}_i(x, y, z)\) is the raw point cloud of the \(i\)-th scan, $\sum$ denotes the geometric merging of multi-scan point clouds. In contrast, the input to the training stage (also the inference stage) when the robot is in motion is:
    \begin{equation}
    \mathbf{X}_{\text{test}} = \mathcal{P} \left( \sum_{i=1}^{N} \mathbf{T}_i \cdot \mathbf{R}_i(x, y, z) \right),
    \end{equation}
where $\mathbf{T}_i$ is the transformation matrix derived from odometry, which introduces alignment errors. This misalignment results in a discrepancy between \(\mathbf{X}_{\text{train}}\) and \(\mathbf{X}_{\text{test}}\), creating a domain gap. To address this, we propose a data augmentation method during training that simulates robot motions. 

Specifically, during scan accumulation, random rotations and translations are applied to each scan to emulate motion-induced inconsistencies. For each scan in a sequence, the augmented input is represented as:  
    \begin{equation}
    \mathbf{X}_{\text{aug}} = \mathcal{P} \left( \sum_{i=1}^{N} \mathbf{T}_{\text{motion}_j} \big( \mathbf{R}_i(x, y, z) \big) + \mathcal{N}(0, \sigma^2) \right),
    \end{equation}
where \(\mathcal{N}(0, \sigma^2)\) represents the Gaussian noise with a mean of 0 and variance \(\sigma^2\), added to simulate sensor noise and motion variability. The transformation \(\mathbf{T}_{\text{motion}_j}\) applies a randomly predefined motion pattern \(j\). To replicate the sparsity typically observed in real scenarios, we employ random scan sampling, where $N$ scans are randomly selected from an initial sequence and then accumulated as input data. Furthermore, non-linear perturbations of odometry are introduced to emulate misaligned poses (e.g., acceleration discontinuities, sharp turns), while noisy IMU signals mimic errors from degraded sensors. By combining these augmentations, the proposed method improves the network's robustness to misaligned, sparse, and noisy inputs, effectively bridging the domain gap between static and dynamic motion conditions.

\subsection{Adaptive Calibration and Densification Network}
\label{sec: 3.3}

    \begin{figure}
        \centering
        \includegraphics[width=0.38\textwidth]{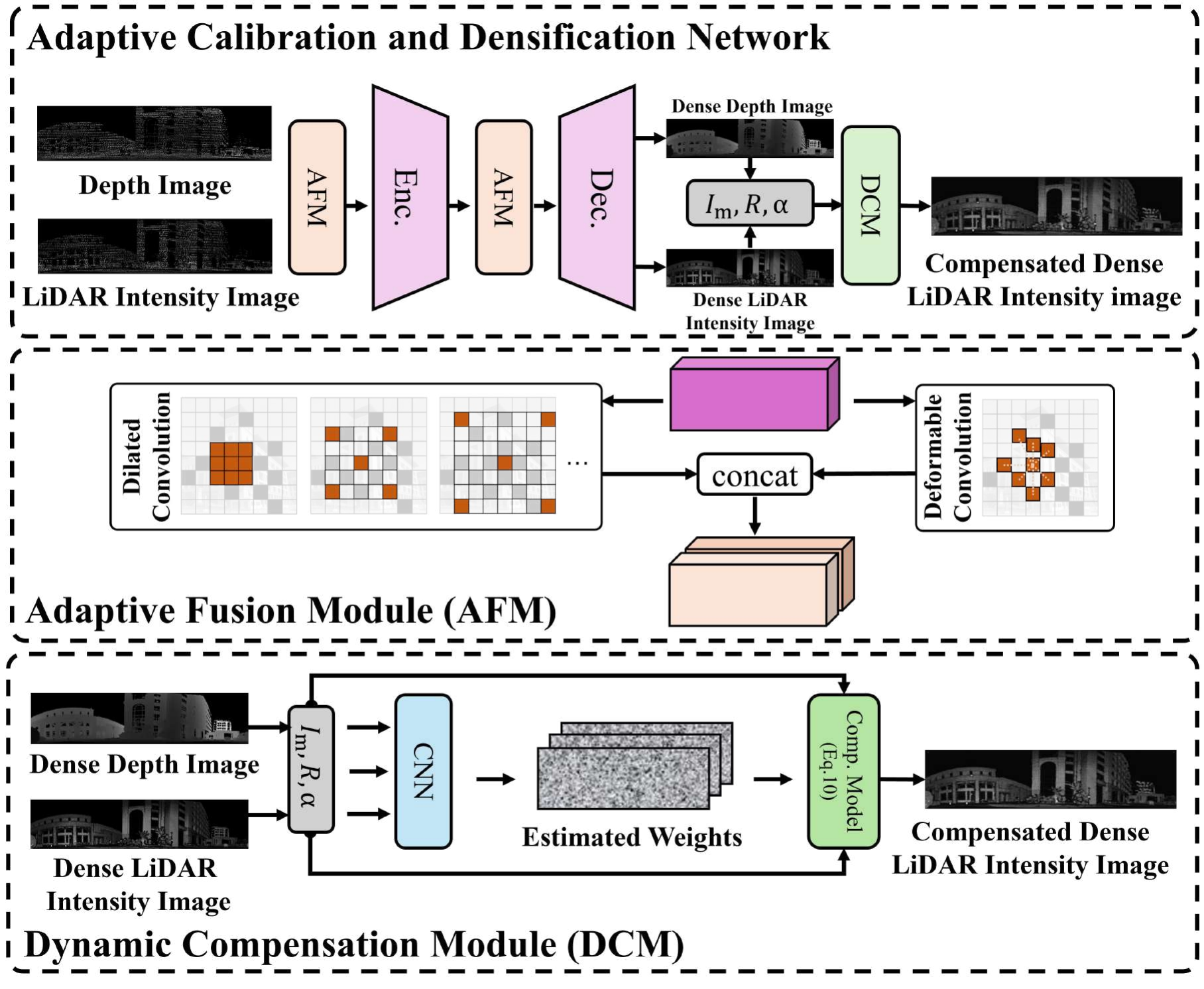}
        \caption{Framework of the proposed method for LiDAR intensity image densification. The framework includes an Adaptive Feature Module (AFM) for feature extraction, an Encoder-Decoder structure for spatial reconstruction, and a Dynamic Compensation Module (DCM) to correct intensity.}
        \label{fig: network}
        \vspace{-10pt}
    \end{figure}

Considering that convolutional neural networks (CNNs) effectively extract spatial features with computational complexity approximately proportional to $\mathcal{O}(n)$ while maintaining efficiency, they were chosen as the backbone of our densification network. In contrast, Transformers \cite{vaswani2017attention} exhibit quadratic complexity $\mathcal{O}(n^2)$, limiting their application in resource-constrained scenarios \cite{zamir2022restormer,liang2021swinir}. Similarly, diffusion models \cite{ho2020denoising} rely on iterative processes, leading to long inference times \cite{li2023diffusion} that hinder real-time applications. Fig. \ref{fig: network} shows our densification network, which employs a U-shaped architecture \cite{ronneberger2015u} with an Adaptive Fusion Module (AFM) and a Dynamic Compensation Module (DCM) to achieve densification. These modules address two key challenges: 1) adaptively fusing multi-scale information from sparse LiDAR points and 2) mitigating intensity decay with distance and incidence angle.

Let $L_{1} \in \mathbb{R}^{H\times W}$ and $D_{1} \in \mathbb{R}^{H\times W}$ be the LiDAR intensity and depth images as input into the network. Depth information is introduced in the intensity image reconstruction during the intensity compensation step. To maximize the utilization of cross-modal interaction and reduce model redundancy, $L_{1}$ and $D_{1}$ are first concatenated along the channel dimension as input to the network. Then, these inputs are processed through an encoder-decoder structure integrated with AFM to produce preliminary reconstructions of dense intensity images $L_{2} \in \mathbb{R}^{H\times W}$ and depth images $D_{2} \in \mathbb{R}^{H\times W}$, where the first and second channels of the decoder are $L_{2}$ and  $D_{2}$, respectively. Furthermore, to eliminate artifacts from the signal intensity decay, DCM is applied to compensate for these effects, resulting in the final dense intensity image $L{3}\in \mathbb{R}^{H\times W}$. DCM employs a learning-based dynamic compensation model to avoid manually designing the parameters required in traditional methods while maintaining end-to-end reconstruction. 

\noindent
\textbf{Encoder and Decoder.}
The encoder comprises four stacked blocks. Each block contains two sequentially connected $3 \times 3$ convolutional layers with ReLU activation, followed by an average pooling layer that downsamples by a factor of 2. Symmetrically, the decoder consists of four blocks. In each decoder block, the input feature map is first upsampled via a transposed convolution module, and then processed by two $3 \times 3$ convolutions for local information fusion. To enhance multi-scale feature integration, we adopt two strategies: 1) inserting an AFM module between the encoder and decoder to facilitate multi-scale spatial information; and 2) following the U-Net architecture \cite{ronneberger2015u}, concatenating corresponding feature maps of identical spatial dimensions from the four encoder blocks into the decoder.

\noindent
\textbf{Adaptive Fusion Module.}
Due to the scanning characteristics of NRS-LiDAR, intensity images captured in short durations are typically sparse and unevenly distributed. Besides, since objects in scenes possess multiscale attributes and vary in distance, fusing raw LiDAR intensity data for incomplete information densification demands that the model possess adaptive capabilities and the ability to fuse multiscale structural features. Traditional convolutional kernels with limited receptive fields struggle to capture information across large regions, and larger kernels significantly increase computational cost. Inspired by\cite{chen2017deeplab, kong2018recurrent}, we utilize parallel dilated convolutions and deformable convolutions to adaptively integrate multiscale information (Fig. \ref{fig: network}). Specifically, parallel dilated convolutions with varying dilation rates expand the receptive field with minimal computational overhead, enhancing multi-scale feature extraction. Simultaneously, deformable convolutions adaptively fuse information based on local contexts, augmenting the model’s ability to extract fine details. Outputs from these parallel convolutions are concatenated along the channel dimension to form the AFM module’s output features.

    \begin{figure}
        \centering
        \includegraphics[width=0.38\textwidth]{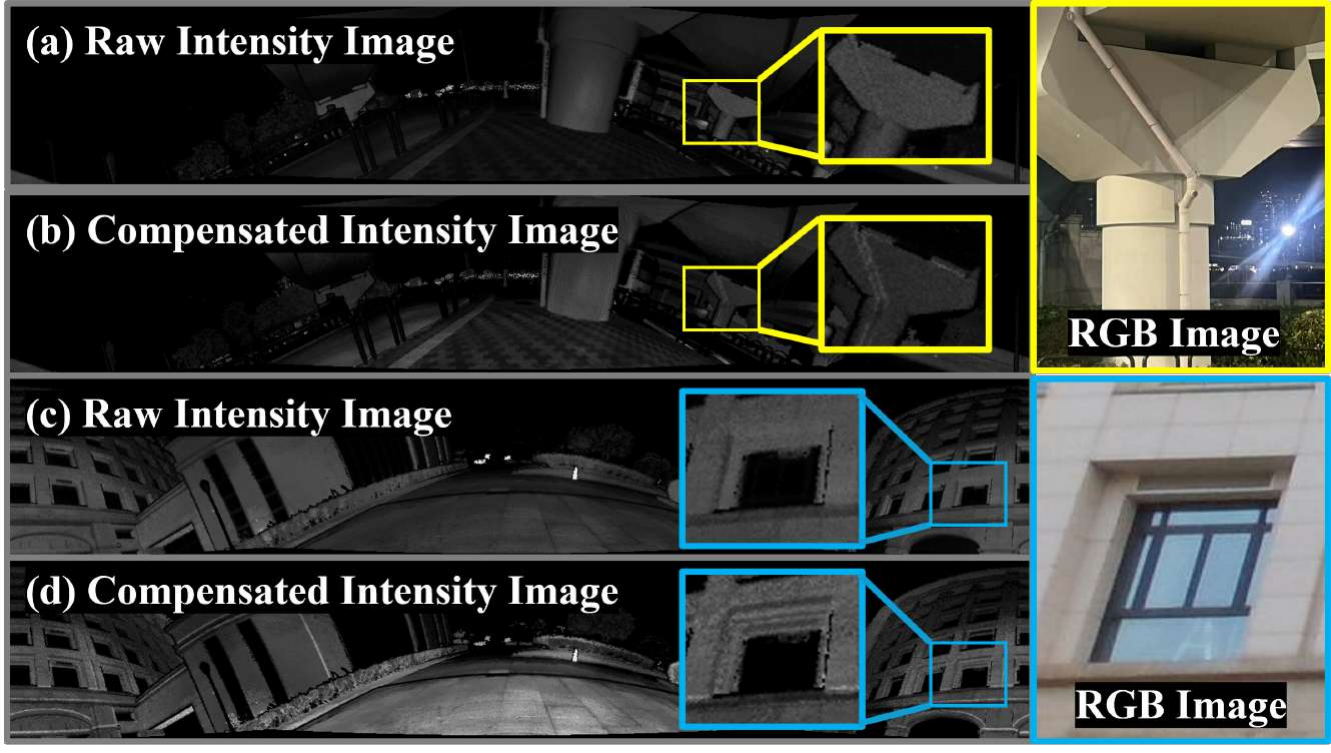}
        \caption{Comparison between the raw dense intensity and the compensated result. The compensated result maintains more consistent intensity values for objects with the same surface material (yellow boxes in  (a) and (b)) and exhibits clearer texture details (\textcolor{blue}{blue} boxes in (c) and (d)).}
        \label{fig:com_compare}
        \vspace{-10pt}
    \end{figure}

\noindent
\textbf{Dynamic Compensation Module.}
The LiDAR intensity is primarily influenced by three factors: the
distance traveled by light, the object's surface reflectance, and the incident angle of light. In general, it is assumed that the reflectance of the object's surface remains relatively consistent. Therefore, compensating for the effects of distance and incident angle is critical for intensity measurement\cite{fang2014intensity}. As introduced in\cite{fang2014intensity}, the captured intensity can be expressed as:
    \begin{equation}
         I(R, \alpha, \rho) \propto P(R, \alpha, \rho) = \eta(R) \frac{I_e \rho \cos \alpha}{R^2},
        \label{eq:theoretical_model}
    \end{equation}
where \(I_e\) represents the laser's emission power, \(R\) denotes the distance from the LiDAR to the object's surface, \(\alpha\) is the incident angle, and \(\rho\) indicates the surface reflectance of the object. The term \(\eta(R)\), which accounts for the near-distance effect, can be expressed as:
    \begin{equation}
        \eta(R) = 1 - \exp\left\{ -\frac{2r_d^2 (R + d)^2}{D^2 S_d^2} \right\},
        \label{etaR}
    \end{equation}
where \(r_d\) is the radius of the laser detector, \(d\) is the offset between the measured range and the true object distance, \(D\) is the lens diameter, and \(S_d\) is the focal length. At near distances, \(\eta(R)\) is small but increases rapidly as \(R\) grows, stabilizing near 1.0 for large \(R\). Rearranging Eq. \eqref{eq:theoretical_model} to express \(I_e\) in terms of other variables:
    \begin{equation}
        I_c \approx I_e = C\frac{I(R, \alpha, \rho) \cdot R^2}{\eta(R) \cdot \rho \cos \alpha},
        \label{eq:Ie}
    \end{equation}
where \(C\) is a proportionality constant that accounts for system-specific calibration factors. \(I_c\) is the corrected or true intensity.

\begin{figure*}
    \centering
    \includegraphics[width=0.72\textwidth]{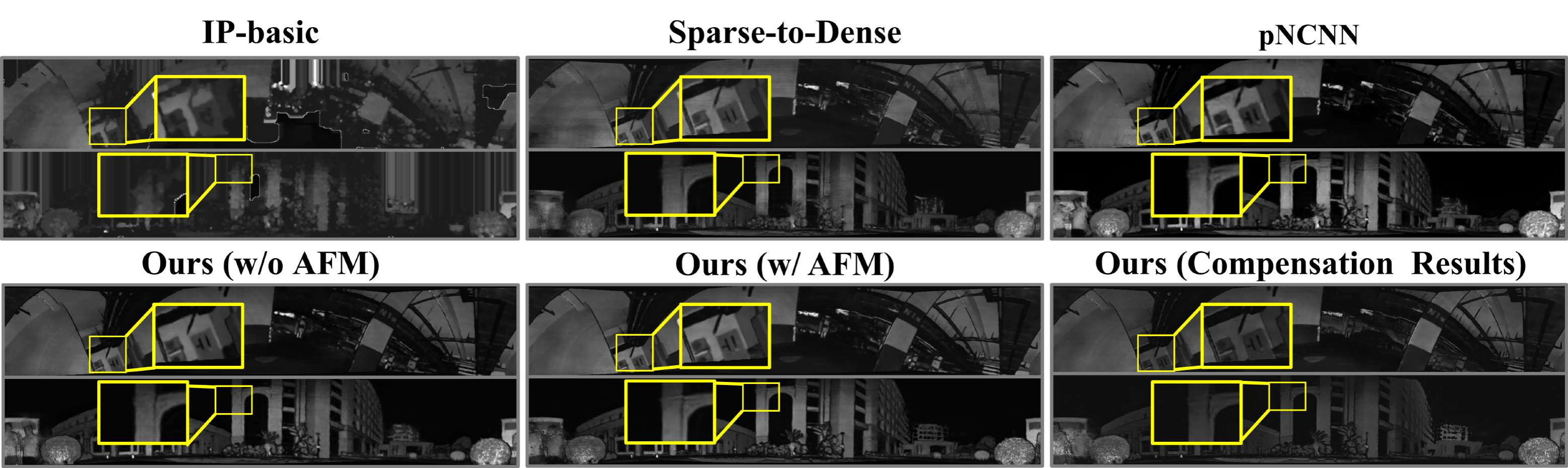}
    \caption{Qualitative comparison of intensity image densification across different methods in indoor and outdoor scenarios. IP-basic \cite{ku2018defense}, a non-learning method, struggles to recover structures in regions with large holes, resulting in distorted outputs. Sparse-to-Dense \cite{ma2019self} and pNCNN \cite{eldesokey2020uncertainty} exhibit noticeable deficiencies in both smoothness and edge preservation. Our method with AFM demonstrates superior performance in preserving edge structures and enhancing details. The compensation results show that the intensity values for the same surface material (e.g., the same wall) are maintained more consistently (\textcolor{yellow}{yellow} boxes).}
    \label{fig:qualitative_comparison}
    \vspace{-10pt}
\end{figure*}

In practice, we collected data for objects with the same material under controlled angles and distances, fitted the required parameters by Eq. \ref{eq:Ie}, and applied them to generate supervised compensated results for training. The compensated results are shown in Fig. \ref{fig:com_compare}. However, the above methods are further limited by data collection constraints and sensor configurations, making them inadequate for capturing complex, non-linear relationships in diverse environments. To address this, we use a simplified, parameterized compensation model, as shown in Eq. \eqref{eq:distance_compensation_learnable} and \eqref{eq:final_compensation_learnable}, which is based on Eq. \eqref{eq:Ie} but incorporates an additional decay factor to comprehensively handle both near-distance and far-distance effects. For distance compensation, the distance correction factor \(\frac{R^2}{\eta(R)}\) becomes significantly large when \(R\) is large. Therefore, we approximated it using a second-order polynomial combined with an exponential decay factor to handle far-distance effects.
    \begin{equation}
        g(R) = \frac{\left(\xi R^2 + \beta R + \gamma\right) \cdot \exp(-\lambda R)}{\eta(R)},
    \label{eq:distance_compensation_learnable}
    \end{equation}
where \(\xi\), \(\beta\), \(\gamma\), and \(\lambda\) are learnable parameters predicted dynamically by the network to capture the non-linear relationship between intensity and range. 

For incidence-angle compensation, the intensity is normalized by \(\cos(\alpha)\), following Lambert's cosine law\cite{kaasalainen2011analysis}. By combining both distance and angle compensation, the corrected intensity \(I_{\text{com}}\) is expressed as:
    \begin{equation}
        I_{com} = \frac{I_m \cdot g(R)}{\cos(\alpha) \cdot \kappa}
        \label{eq:final_compensation_learnable}
    \end{equation}
where \(I_{\text{m}}\) represents the measured intensity, equal to the $I(R, \alpha, \rho)$ in Eq.6, and \(\kappa\) is a learnable parameter to account for deviations from ideal Lambertian reflectance and sensor-specific factors. In our method, we employ a lightweight CNN comprising three convolutional layers with \(3 \times 3\) kernels to dynamically predict per-pixel compensation parameters  (\(\xi\), \(\beta\), \(\gamma\), \(\lambda\), \(\kappa\)) from local context. The propagation distance \(R\) is obtained from the densification module’s output depth map \(D_2\). To compute the incidence angle \(\alpha\), we first back-project \(D_2\) into 3D to recover surface points, estimate per-pixel surface normals via local PCA plane fitting, and then use the LiDAR beam direction \(\mathbf{d}\) (from the sensor to each 3D point) to derive \(\cos(\alpha) = |\mathbf{d} \cdot \mathbf{n}|\), where \(\mathbf{n}\) is the unit normal. These parameters are then applied to \(I_{\text{m}}\) using Eq. \eqref{eq:final_compensation_learnable}, enabling the model to perform intensity compensation. To train the CNN module in the DCM, we use a dataset of 300 manually selected, well-calibrated compensated images as supervision data, optimizing the network with an $L_1$ loss function. We specifically selected samples with complete scene structures and clear textures while filtering out inputs containing excessive clutter (e.g., vegetation) or dynamic artifacts. The input to the CNN includes the original densified intensity image, as well as the per-pixel distance and incidence angle, allowing the CNN to dynamically predict per-pixel compensation parameters.


\section{Experiment}

We conducted a series of experiments to evaluate the effectiveness and robustness of the proposed method, focusing on the real-time LiDAR intensity image densification and its applications in loop closure detection and traffic lane detection. Data collection and real-world experiments were conducted using a Livox MID-360, mounted on a wheeled ground robot equipped with an NVIDIA RTX 3090 GPU and an Intel i7-1165G7 CPU. The experiments covered diverse scenarios, including varying lighting conditions (daylight, overcast, nighttime) and urban environments, to comprehensively evaluate the proposed method.

\subsection{Evaluation of LiDAR Intensity Image Densification}
\label{sec: net_exp}

    \begin{figure}
        \centering
        \includegraphics[width=0.38\textwidth]{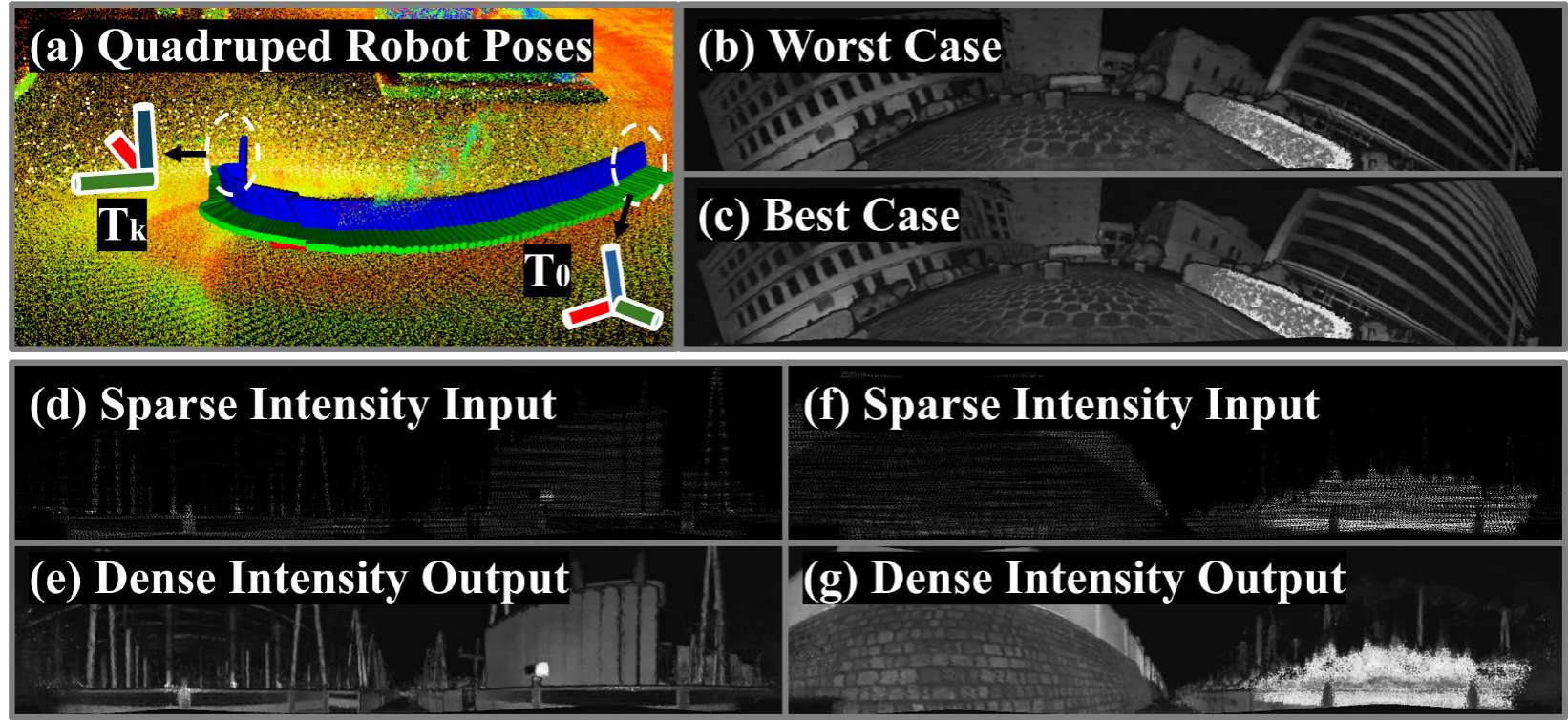}
        \caption{(a)-(c) Densification results using 5-frame data with aggressive quadruped robot motion poses on uneven terrain, where (b) shows the worst case (PSNR = 25.75) and (c) shows the best case (PSNR = 26.82); (d)-(g) input and intensity image densification outputs on the public dataset\cite{wang2024sfpnet}.}
        \label{fig:extend_exp}
        \vspace{-10pt}
    \end{figure}

    \begin{table}[t!]
    \centering
    \caption{Quantitative Comparison of Densification Performance}
    \label{table:intensity_comparison}
    \scriptsize 
    \setlength{\tabcolsep}{4pt} 
    \renewcommand{\arraystretch}{1.2} 
    \begin{tabular}{@{}lcccc@{}}
    \toprule
    \textbf{Method} & \textbf{PSNR / SSIM} $\uparrow$ & \textbf{RMSE} $\downarrow$ & \textbf{MAE} $\downarrow$ & \textbf{Time (ms)} $\downarrow$ \\
    \midrule
    IP-basic\cite{ku2018defense} & 22.448 / 0.642 & 0.0672 & 0.0471 & 10.6 (CPU) \\
    pNCNN\cite{eldesokey2020uncertainty} & 25.746 / 0.718 & 0.0537 & 0.0282 & 168 (GPU) \\
    Sparse-to-Dense\cite{ma2019self} & 26.277 / 0.724 & 0.0505 & 0.0272 & 69 (GPU) \\
    LiDAR-Net\cite{dai2022lidar}&24.643/0.703 & 0.0585 & 0.0311 & 105 (GPU)\\
    Marigold-dc\cite{viola2024marigold} & 24.21 / 0.691 & 0.613 & 0.354 & 26140 (GPU) \\
    Ours (w/o AFM) & 26.051 / 0.721 & 0.0521 & 0.0276 & \textbf{36} (GPU) \\
    Ours (w/ AFM) & \textbf{27.152 / 0.767} & \textbf{0.0492} & \textbf{0.0247} & 52 (GPU) \\
    \bottomrule
    \end{tabular}
    \vspace{-10pt}
    \end{table}

    \begin{figure*}
        \centering
        \includegraphics[width=0.81\textwidth]{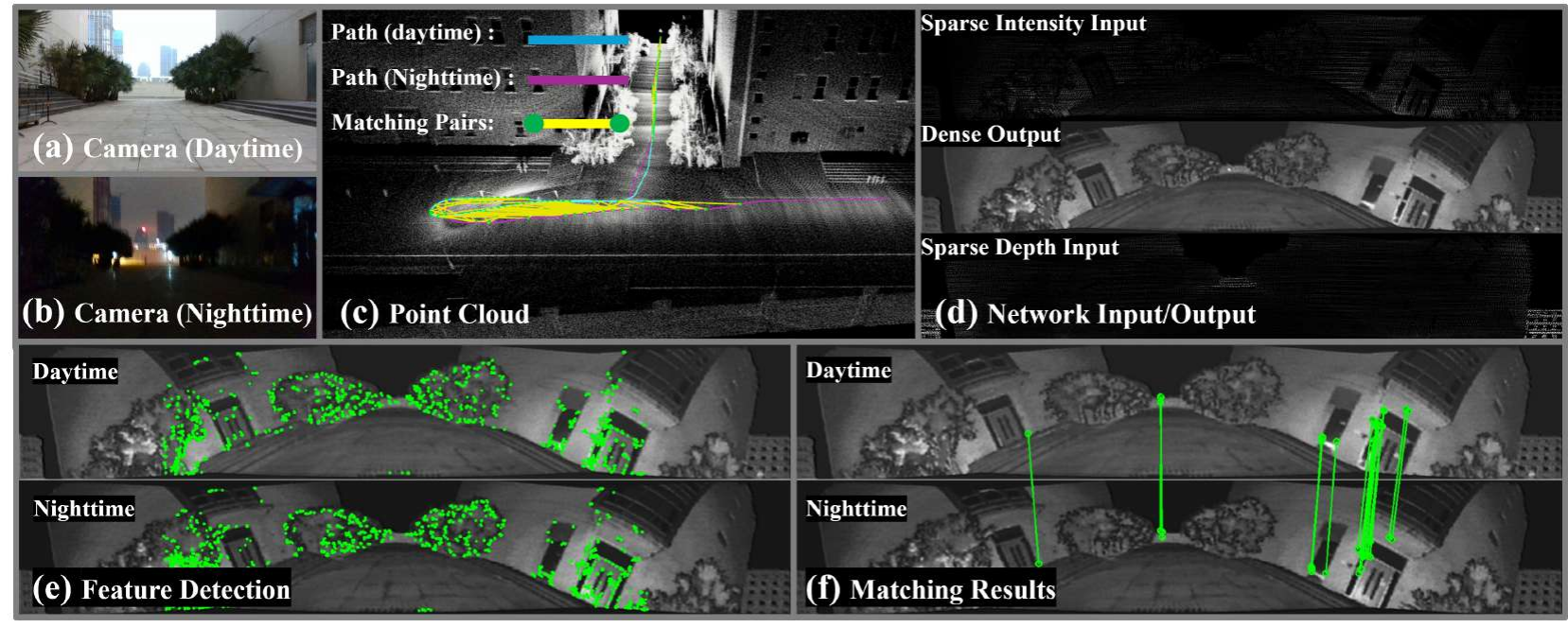}
        \caption{Loop closure results across different times, with one dataset collected during daytime and the other at night (panoramic spherical projection). Our method achieves robust, real-time feature matching via visual loop closure detection, showing consistent performance under significant lighting variations.}
        \label{fig:Place re-id}
        \vspace{-10pt}
    \end{figure*}

    \begin{table}[t!]
        \centering
        \caption{Quantitative Results of Different Loop Closure Detection Methods in Different Environments}
        \label{tab:loop}
        \setlength{\tabcolsep}{4pt} 
        \renewcommand{\arraystretch}{1.2} 
        \begin{tabular}{llccc}
            \toprule
            \textbf{Env.} & \textbf{Method} & \textbf{Detected Loops} & \textbf{True Positives} & \textbf{False Positives} \\
            \midrule
            \multirow{3}{*}{Env. 1} 
            & ISC\cite{wang2020intensity} & 53 & 32 (60\%) & 21 (40\%) \\
            & IRIS\cite{wang2020lidar} & 36 & 19 (53\%) & 17 (47\%) \\
            & Ours & 73 & 70 (96\%) & 3 (4\%) \\
            \midrule
            \multirow{3}{*}{Env. 2} 
            & ISC\cite{wang2020intensity} & 88 & 59 (67\%) & 29 (33\%) \\
            & IRIS\cite{wang2020lidar} & 96 & 72 (75\%) & 24 (25\%) \\
            & Ours & 138 & 130 (98\%) & 8 (2\%) \\
            \midrule
            \multirow{3}{*}{Env. 3} 
            & ISC\cite{wang2020intensity} & 152 & 133 (88\%) & 19 (12\%) \\
            & IRIS\cite{wang2020lidar} & 162 & 78 (48\%) & 84 (52\%) \\
            & Ours & 189 & 175 (93\%) & 14 (7\%) \\
            \bottomrule
        \end{tabular}
        \vspace{-10pt}
    \end{table}
    
To assess the performance of our proposed densification network, we conducted comparative and ablation experiments. Comparative experiments evaluated the quality of densified images generated by our method against CNN-based and traditional methods, both qualitatively and quantitatively. For qualitative evaluation, we visually compared the generated images with the results of competing methods (see Fig. \ref{fig:qualitative_comparison}). Quantitatively, we used standard metrics such as Peak Signal-to-Noise Ratio (PSNR), Structural Similarity Index Measure (SSIM), Root Mean Squared Error (RMSE), and Mean Absolute Error (MAE) to measure accuracy while also evaluating inference time to assess computational efficiency.

Since there are limited public datasets or prior works specifically addressing LiDAR intensity image densification, we trained several depth completion models, included traditional interpolation methods for comparison\cite{ma2019self, eldesokey2020uncertainty, ku2018defense,viola2024marigold}, and reproduced LiDAR-Net \cite{dai2022lidar}, a method specifically designed for LiDAR intensity image completion, providing only quantitative results due to its non-open-source nature. An ablation study was conducted to analyze the contribution of the Adaptive Fusion Module (AFM). Table \ref{table:intensity_comparison} presents the quantitative results, while Fig. \ref{fig:qualitative_comparison} demonstrates the qualitative superiority of our method. Our method achieves high accuracy (highest PSNR/SSIM, lowest RMSE/MAE) while maintaining competitive runtime. Notably, as Marigold-dc\cite{viola2024marigold} is a zero-shot diffusion-based method designed for depth completion, its performance is less effective in our task context; therefore, we only provide its quantitative results.

To evaluate the performance of our method in generating dense intensity images during robot motion, we tested on data (5-frame accumulation) captured when robot was subject to aggressive motion poses on uneven road. The resulting PSNR values (25.75–26.82) demonstrate the robustness of our approach against alignment noise and non-linear motion artifacts (Fig. \ref{fig:extend_exp}). Additionally, we compared real-time results at various robot speeds in the supplementary material.

Although we wish to qualitatively evaluate the generalization ability of our method on the other datasets, to the best of our knowledge, we have not found such public datasets with dense ground truth. However, we show qualitative visualizations on a public dataset \cite{wang2024sfpnet} (without dense ground truth), which can further validate the generalization ability of our approach, as illustrated in Fig. \ref{fig:extend_exp}.

    \begin{figure}
        \centering
        \includegraphics[width=0.38\textwidth]{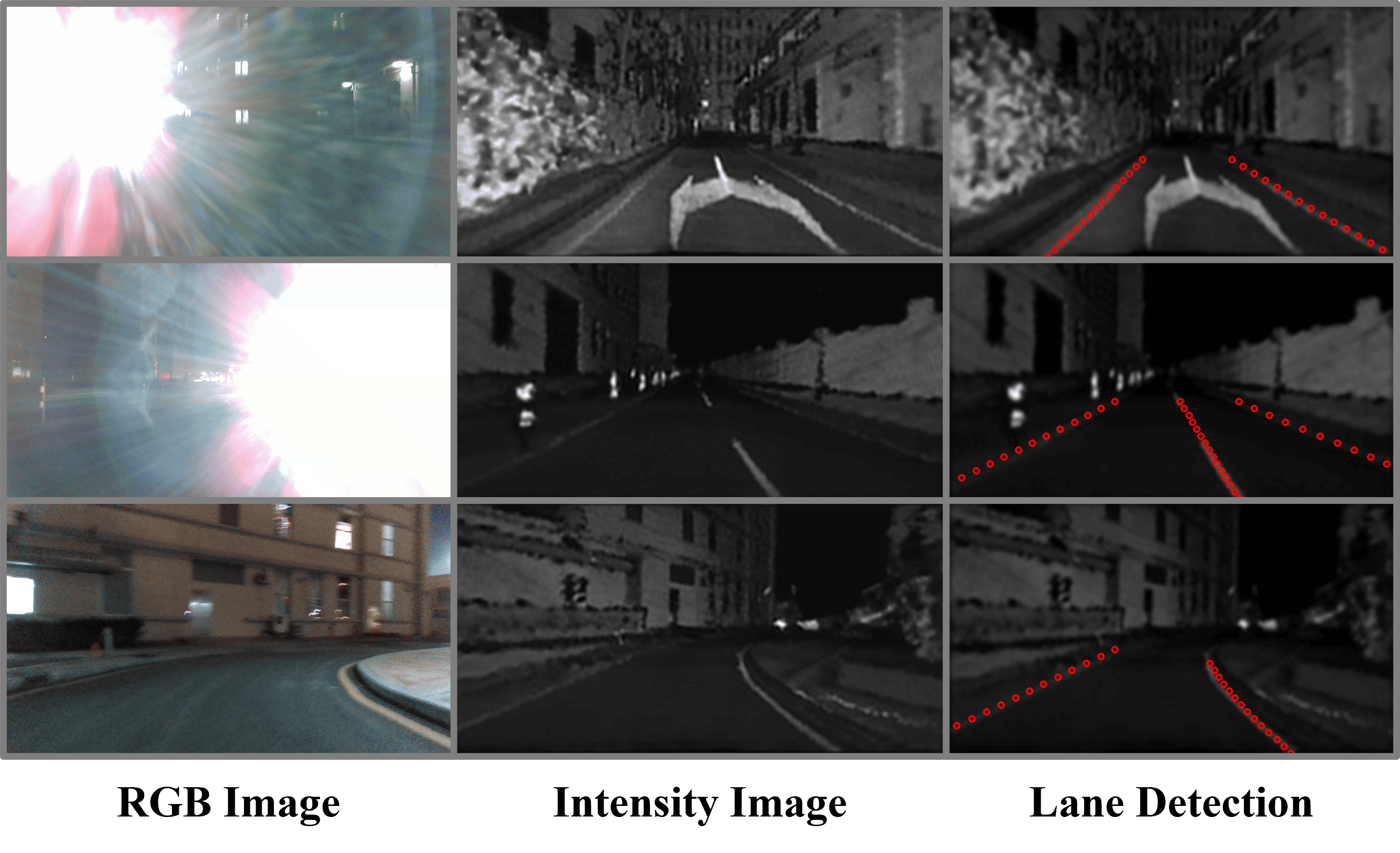}
        \caption{Top: Single-lane scene with light pollution. Middle: Multi-lane scene with illumination from oncoming vehicles. Bottom: A curved road under insufficient lighting conditions.}
        \label{fig:lane-detection-corner-cases}
        \vspace{-10pt}
    \end{figure}

\subsection{Evaluation in Specific Applications}
\label{sec: task_exp}

\subsubsection{Loop Closure Detection}
To validate the effectiveness of our method, we adopt the approach of\cite{shan2021robust} but replace the high-resolution intensity data acquired by high-cost LiDAR with the dense intensity images generated by our method using Livox MID-360. Specifically, \cite{shan2021robust} extracts ORB feature descriptors from intensity images for loop closure detection. These descriptors are then represented as bag-of-words vectors for querying loop closures. After retrieving a candidate match, ORB descriptor matching is performed, and outliers are removed using PnP RANSAC\cite{fischler1981random} for verification. 
We conducted experiments in three distinct environments under two lighting conditions: daytime and nighttime (Fig. \ref{fig:Place re-id}), using data collected under varying conditions to assess the generalization capability of our approach. Because the adopted loop closure detection method\cite{shan2021robust} is not based on deep learning, for a fair comparison, we compared our loop closure detection results with other non-deep-learning methods, including \cite{wang2020lidar, wang2020intensity}, which are representative LiDAR-descriptor based methods. The results in Table \ref{tab:loop} demonstrate that our method consistently outperforms other approaches, maintaining reliable performance despite changes in illumination. 

\subsubsection{Traffic Lane Detection}
LiDAR intensity images reveal a strong contrast between traffic lane markings and the road surface due to differences in surface material physical properties. Leveraging dense intensity images generated by our method, we achieved accurate lane detection under diverse conditions, including challenging scenarios for cameras (Fig. \ref{fig:lane-detection-corner-cases}). Lane detection was performed with LaneATT \cite{tabelini2021keep} as a baseline model. The model was trained on the TuSimple dataset \cite{tusimple-lane-detection} and fine-tuned on our dataset, which consists of 1,300 training images, 300 validation images, and 200 test images, with a maximum of four lanes per image. A ResNet-34 backbone \cite{he2016deep} was used for fine-tuning the model. The detection performance was evaluated across three distinct environments with varying lighting conditions. The overall detection accuracy ranged from 93.3\% to 96.1\%, while the false detection rate (FDR) and false negative rate (FNR) remained low, with averages of 6.08\% and 6.22\%, respectively. The results show the robustness of our approach, which achieves stable and accurate lane detection, particularly under adverse lighting conditions (Fig. \ref{fig:lane-detection-corner-cases}), outperforming camera-based methods.


\section{Conclusion} 
In this work, we introduce a novel benchmark dataset and a framework for generating dense LiDAR intensity images from sparse measurements, addressing key challenges such as domain adaptation, intensity calibration, and intensity images restoration.  Future work will explore generalizing our approach to repeating-scanning LiDAR sensors and balancing real-time performance with densification quality. We also aim to leverage super-resolution LiDAR intensity images to enhance robotic perception in challenging environments.


\bibliographystyle{IEEEtran}
\bibliography{main}





\clearpage
\twocolumn[ 
\begin{center}
    {\LARGE \textbf{Super LiDAR Intensity for Robotic Perception}\\[1em]{ Supplementary Material}} 
\end{center}
\vspace{2em} 
] 


\section{LiDAR Reflectance Compensation}
\begin{figure}[!ht]
    \centering
    \includegraphics[width=0.8\linewidth]{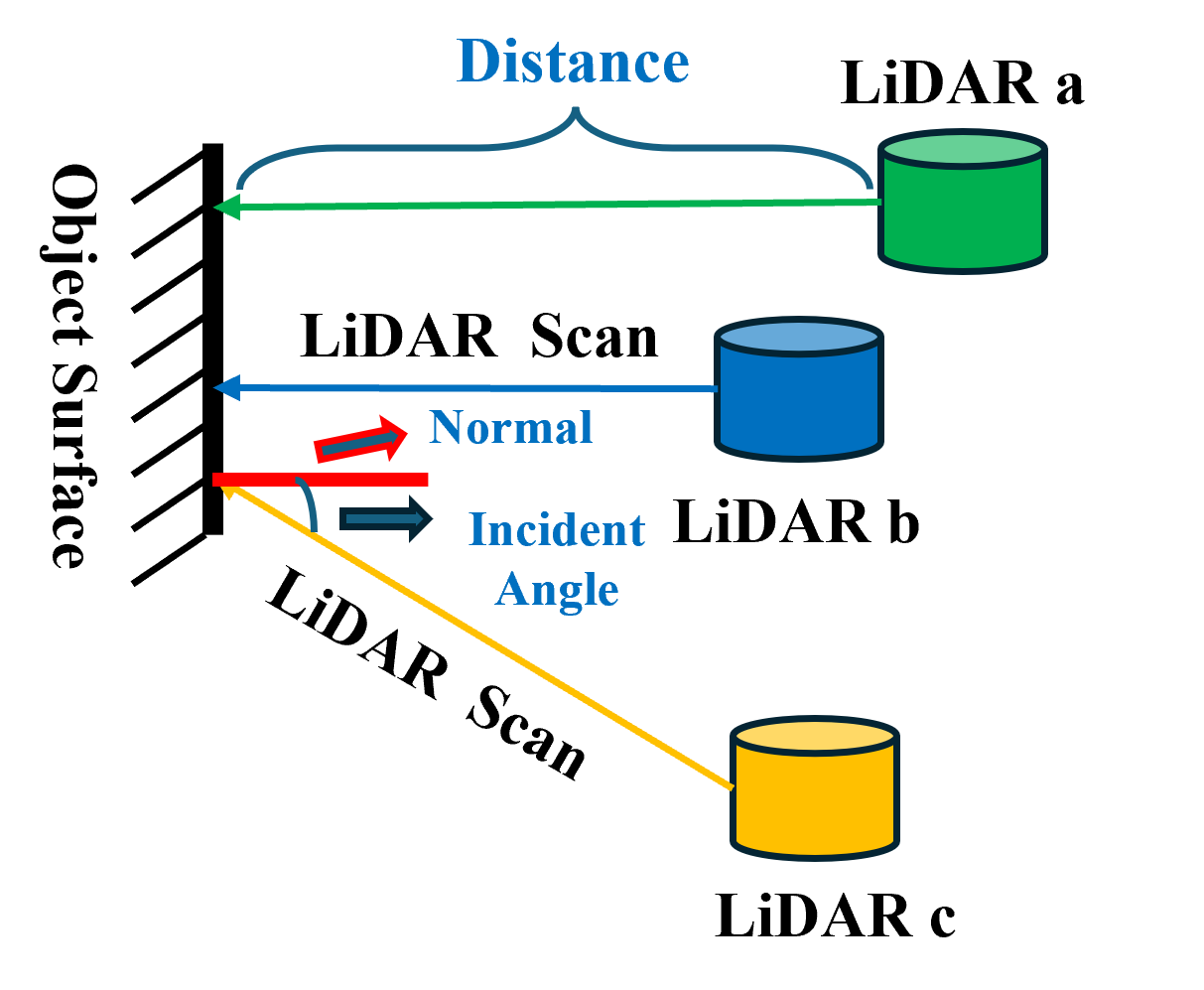}
    \caption{For the same object, LiDAR reflectance differs between a and b due to distance, and between a and c due to angle.}
    \label{fig:com}
\end{figure}
\begin{figure}[h]
    \centering
    \includegraphics[width=\linewidth]{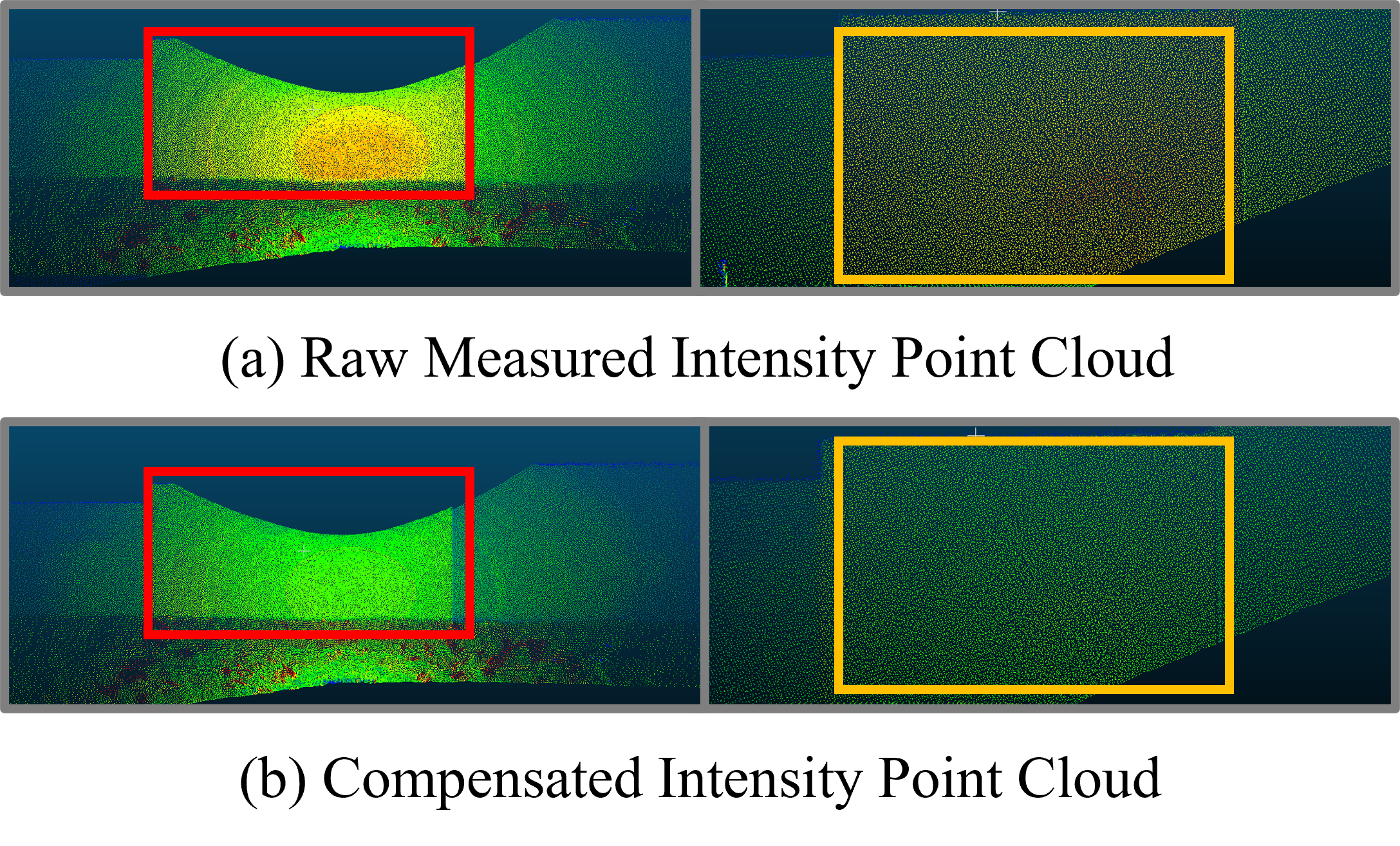}
    \caption{For the same object (red and orange boxes), raw measurements show inconsistencies in reflectance due to variations in angle and distance (color represents reflectance values). After compensation, the reflectance for the same object in the point cloud is maintained within a consistent range.}
    \label{fig:com_wall}
\end{figure}
\begin{figure}[h]
    \centering
    \includegraphics[width=\linewidth]{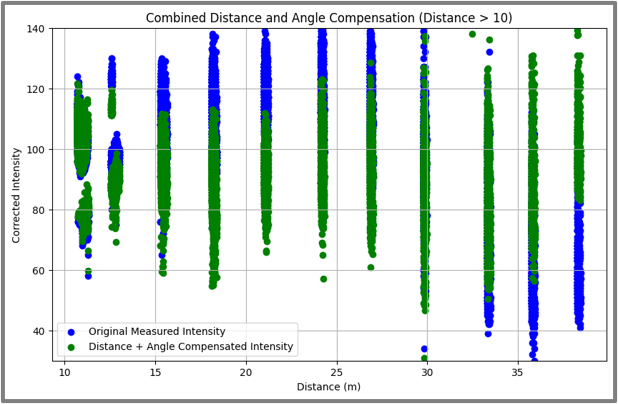}
    \caption{Reflectance values distributed across different distances. Blue points represent the original measured intensity, showing significant variation with distance, while green points represent the values after applying our compensation model, maintaining a more consistent range.}
    \label{fig:com_chart}
\end{figure}

As shown in Fig.~\ref{fig:com}, for the same object, the reflectance obtained by the LiDAR beam varies due to differences in distance or incidence angle. In practice, multiple datasets are collected for objects of the same material, where distance and angle are controlled as single variables to fit their respective relationships with reflectance. The model parameters are then obtained based on the formula in the text. An example of the calibration results is shown in Fig.~\ref{fig:com_wall}, after applying our compensation method, the reflectance for the same object remain consistent (see Fig.~\ref{fig:com_wall} (b)), unlike in Fig.~\ref{fig:com_wall} (a), where the reflectance varies with incidence angle or distance. 
Furthermore, to validate the effectiveness of our compensation model at long distances (greater than 10m), we analyzed the intensity distribution of the same object across different distances and angles, along with the corrected results, as shown in Fig.~\ref{fig:com_chart}. The results demonstrate that our compensation method effectively maintains the reflectance values within a consistent range across varying distances and angles for the same object.

\section{Supplementary Experiments}
\subsection{Test on Pubic Dataset}
\begin{figure}
    \centering
    \includegraphics[width=\linewidth]{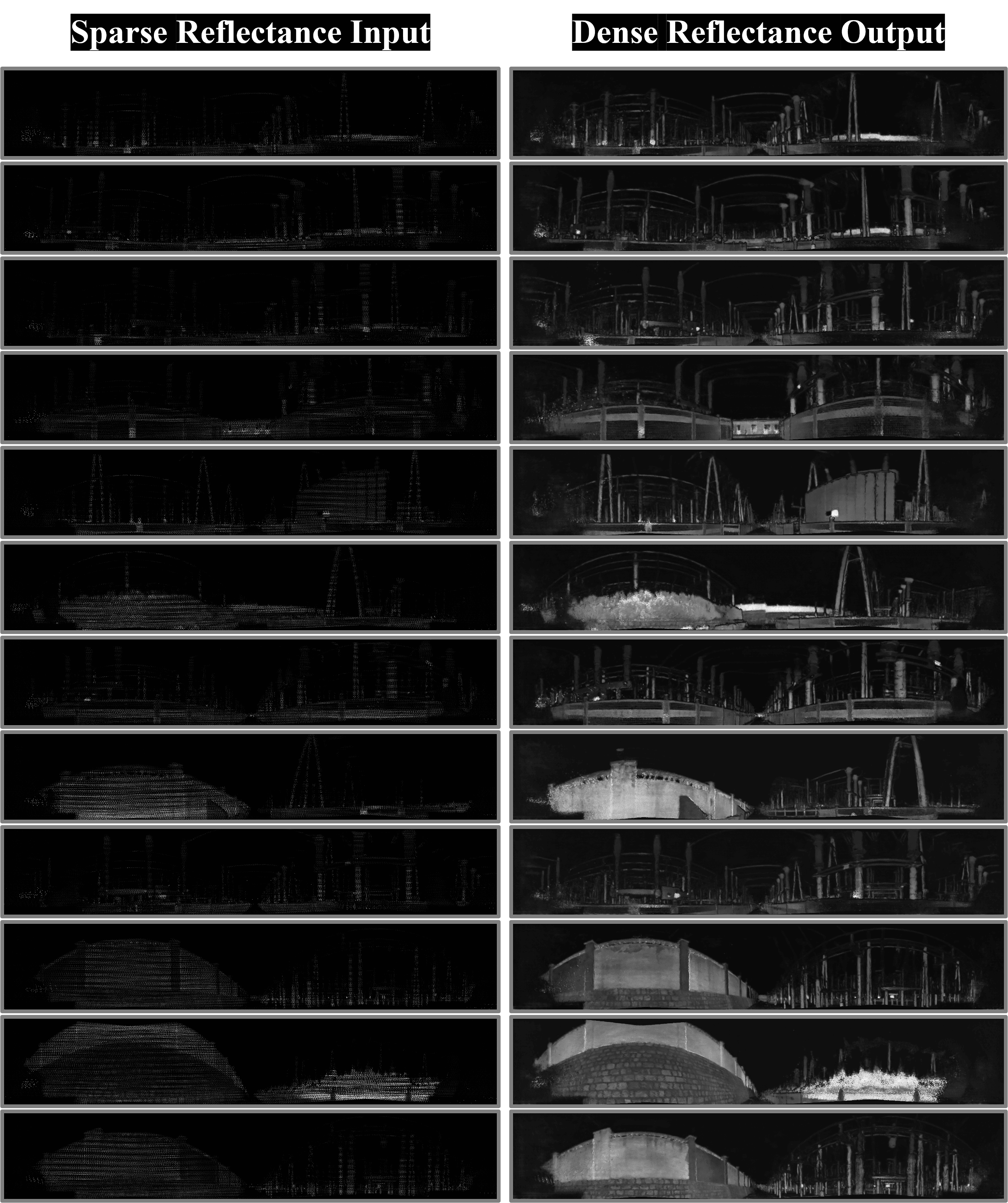}
    \caption{Input and intensity image densification results on the public dataset \cite{wang2024sfpnet}}
    \label{fig:sfpnet}
\end{figure}

As mentioned in Section~\ref{sec: net_exp}, we provide more qualitative visualizations results on a public dataset \cite{wang2024sfpnet} (without dense ground truth), which can further validate the generalization ability of our approach, as illustrated in Fig. \ref{fig:sfpnet}.
\subsection{Impact of Frame Accumulation on Densification Quality}
\begin{figure*}
    \centering
    \includegraphics[width=0.9\linewidth]{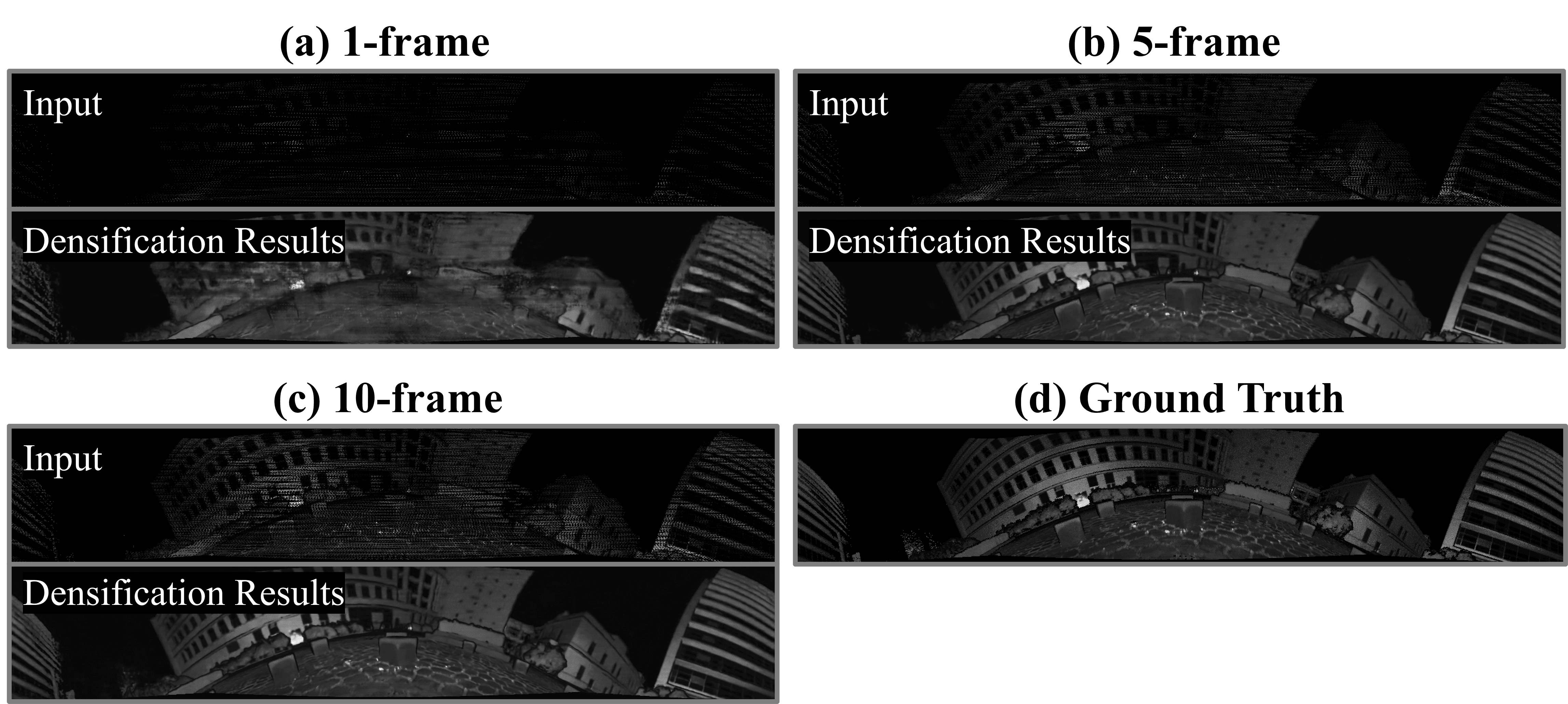}
    \caption{Comparison of dense reflectance image generated from 1-frame, 5-frame, and 10-frame accumulated input}
    \label{fig:frame_compare}
    \vspace{-0.5cm}
\end{figure*}

As mentioned in Section~\ref{limit}, to justify the rationale behind using 5-frame accumulated data as input for our method, we qualitatively and quantitatively compared the densification results by using the single-frame and 10-frame accumulated data as input. As shown in Table~\ref{tab:frame_compare} and Fig.~\ref{fig:frame_compare}, the densification performance of 5-frame input is comparable to that of 10-frame input, while the reconstruction quality of single-frame input is significantly worse. This is because single-frame input is often excessively sparse, resulting in objects lacking sufficient information for accurate reconstruction. It is worth noting that the evaluation here is based on the densification quality of statically collected data. Although the 10-frame input achieves better metric performance, it overlooks the impact of odometry errors and moving objects in the scene during the robot's motion. The input data and the generated images are illustrated in Fig.~\ref{fig:frame_compare}.

\begin{table}[htbp]
\centering
\caption{Quantitative comparison of dense reflectance images generated from 1-frame, 5-frame, and 10-frame accumulated input using different metrics}
\label{tab:frame_compare}
\begin{tabular}{@{}lcccc@{}}
\toprule
\textbf{Frames Input} & \textbf{PSNR} & \textbf{SSIM} & \textbf{RMSE} & \textbf{MAE} \\ \midrule
1-frame & 22.303  &   0.350   &   0.0802       &    0.058   \\
5-frame   &  27.142   &   0.757   &  0.0503          &    0.0263       \\
10-frame       &     27.239      &    0.762           &  0.0496    &     0.0256   \\ \bottomrule
\end{tabular}
\end{table}

\subsection{Impact of Robot Speed on Densification Performance}
\label{app_1}
\begin{figure*}
    \centering
    \includegraphics[width=0.9\linewidth]{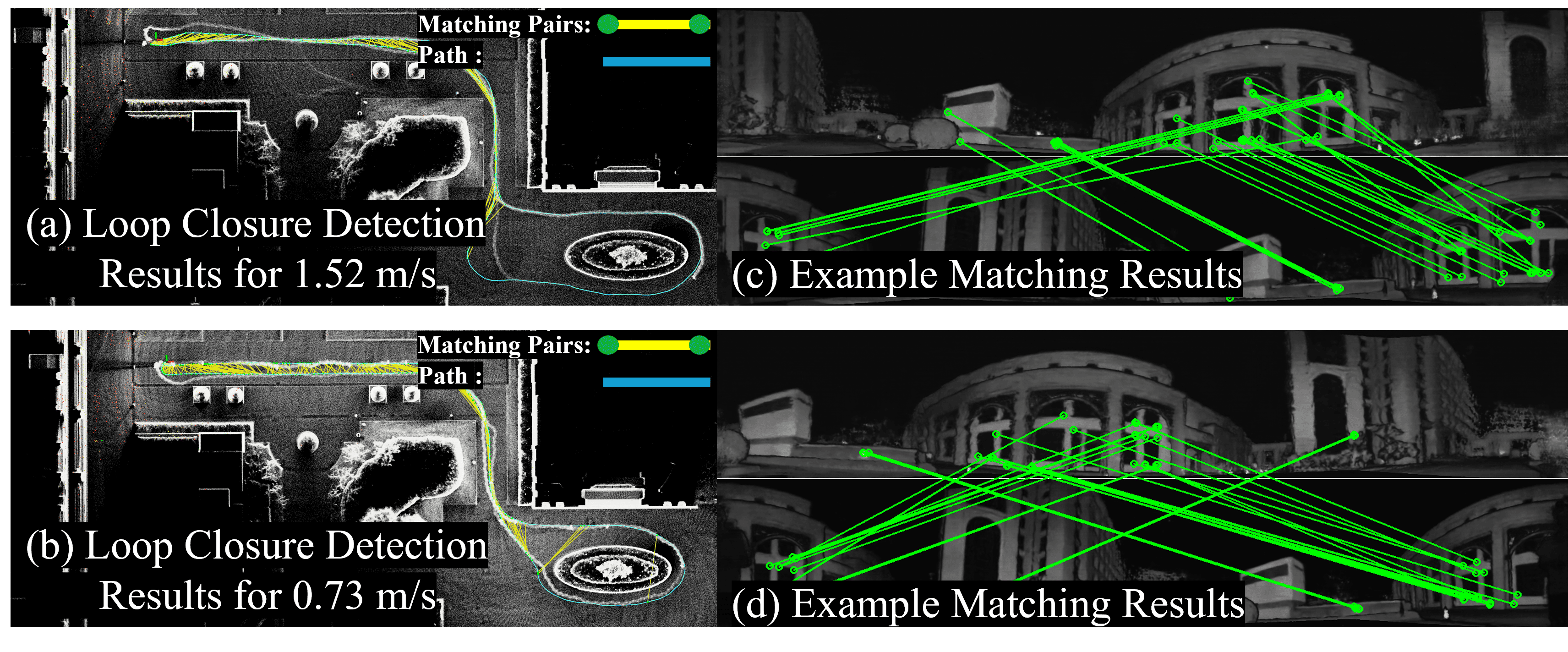}
    \caption{Loop closure detection and feature point matching results at different robot speeds in the same scene.}
    \label{fig:speed}
\end{figure*}

Considering that our method is designed to serve the perception tasks of robots during motion, we conducted experiments to evaluate the dynamic densification results at different robot speeds in the same scenarios, using loop closure detection as the test task. The quantitative comparison results are shown in Table~\ref{table:speed}. Two visual loop closure detection results, as well as the generated image quality and feature point matching results at different speeds, are illustrated in Fig.~\ref{fig:speed}.

\begin{table}[h!]
\centering
\caption{Loop Closure Detection Results Under Different Speeds}
\label{table:speed}
\begin{tabular}{lcc}
\toprule
\textbf{Average Speed} & \textbf{Detected Loops} & \textbf{Trajectory Length} \\ 
  (m/s) &    &  (m) \\ 
\midrule
0.73 & 213 & 193.73 \\
1.52 & 185 & 203.51 \\
\bottomrule
\end{tabular}
\end{table}


\begin{figure*}[h]
    \centering
    \includegraphics[width=0.9\linewidth]{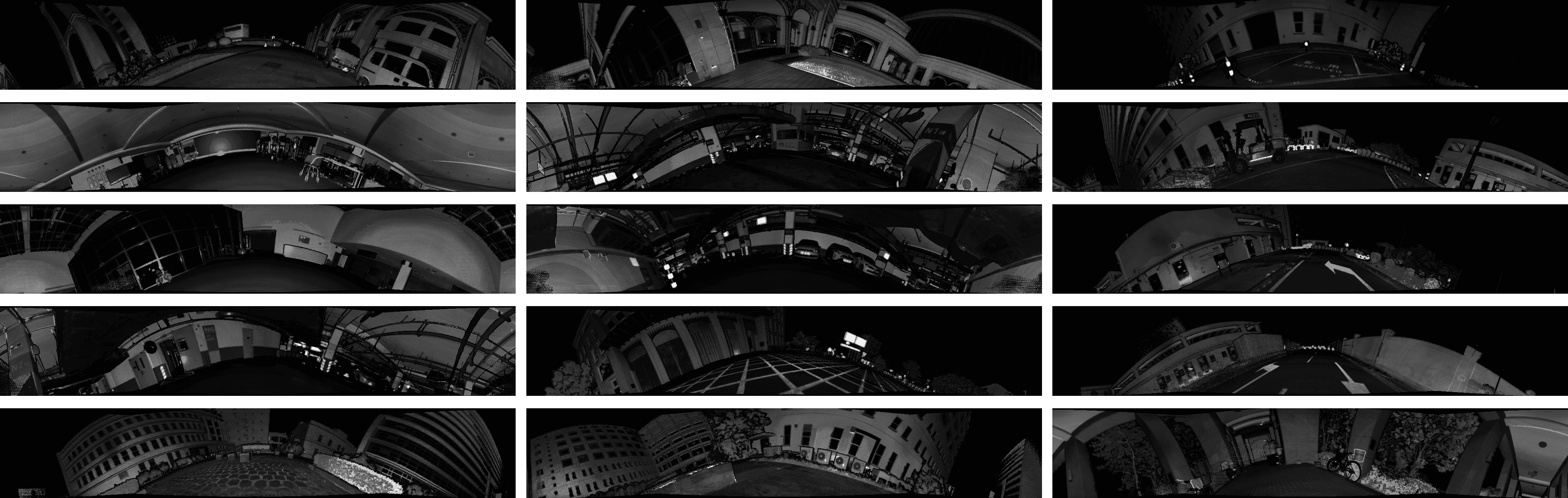}
    \caption{Panoramic reflectance images generated using panoramic projection.}
    \label{fig:range_dataset}
\end{figure*}

\begin{figure*}[h]
    \centering
    \includegraphics[width=\linewidth]{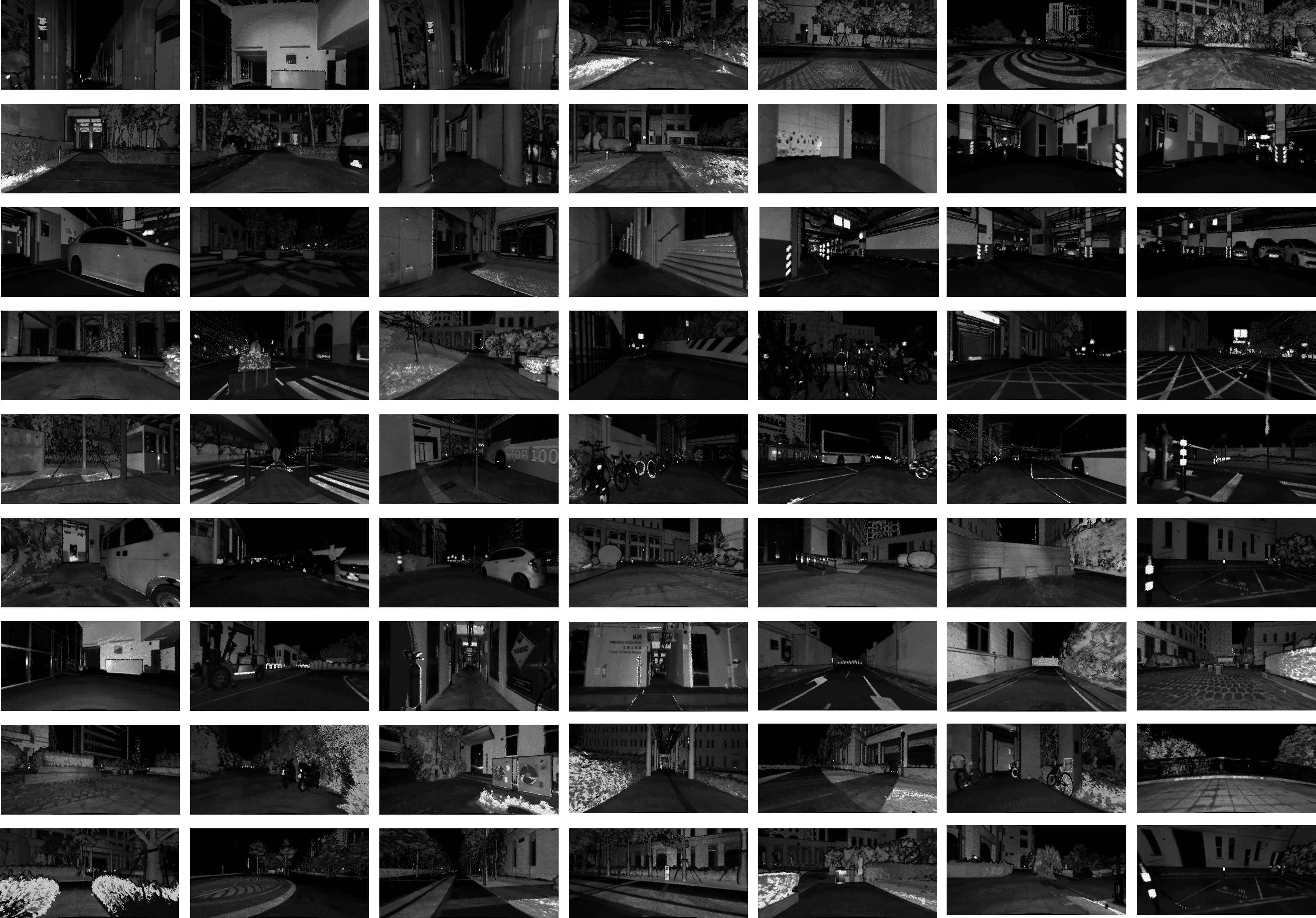}
    \caption{Virtual-camera reflectance images generated using virtual-camera projection.}
    \label{fig:four_dataset}
\end{figure*}
\section{Dataset Construction}

We constructed the Super LiDAR Reflectance dataset, which includes 1,000 raw ROS packets and an additional 5,000 sets of reflectance images with configurations ranging from sparse (1 scan) to dense (500 scans). Specifically, from the 1,000 raw ROS packets, 1,000 panoramic reflectance images were generated using the panoramic mode projection, with some exemplar results shown in Fig.~\ref{fig:range_dataset}. These images cover the full field of view (FoV) of the LiDAR sensor. Similarly, 4,000 sets of reflectance images were generated from the same 1,000 raw ROS packets using the virtual-camera projection mode, with exemplar images shown in Fig.~\ref{fig:four_dataset}. 

Furthermore, the dataset contains two subsets tailored for practical tasks. The first subset is designed to evaluate the performance of generated panoramic dense reflectance images in loop closure detection. The second subset assesses the utility of generated virtual-camera dense reflectance images for traffic lane detection. Both subsets were collected while the robot was in motion, aiming to validate the effectiveness of dense reflectance images in enhancing perception during robot movement.


\section{Limitation and Future Work}
\label{limit}
While the proposed method demonstrates significant advancements in generating dense LiDAR reflectance images, several limitations remain. First, the current approach is not easily generalizable to all LiDAR types due to differences in scanning patterns, which limits its adaptability across various hardware setups. Future work will investigate the generalization and adaptability of the method to LiDAR with different scanning modes to address this limitation. 

Second, the use of single-frame input helps mitigate the static-to-dynamic domain gap issue but results in excessively sparse data, which can impact performance. Addressing this trade-off is an important direction for future work to further enhance the method's robustness and performance, particularly in dynamic and challenging environments.

In addition, this paper presents a novel method for generating dense LiDAR reflectance images from sparse input, addressing challenges such as domain adaptation, intensity calibration, and restoration. In the future, we will leverage the generated super LiDAR reflectance images to support a wider range of robotic perception tasks, enabling their use in dynamic and complex environments.

Moreover, the integration of camera images with LiDAR reflectance images will be explored. By coloring reflectance images with camera information, the densified images can provide more detailed semantic understanding, bridging the gap between LiDAR and camera domains. This will open up new possibilities for vision-based tasks, such as semantic segmentation, where camera-based methods can be adapted to operate on LiDAR data. These advancements will further expand the applicability of the proposed method in diverse robotic perception scenarios. 



\end{document}